%% file: main.tex
\documentclass{article}

\usepackage[final]{corl_2025} 

\usepackage{lipsum}
\usepackage{wrapfig}
\usepackage{multirow}
\usepackage{booktabs}
\usepackage{rotating}

\usepackage{amsfonts}
\usepackage{amsmath}
\usepackage{wrapfig}
\usepackage{graphicx}
\usepackage{lipsum} 
\usepackage{wrapfig,booktabs}

\usepackage{bbm}

\title{ScrewSplat: An End-to-End Method for \\ Articulated Object Recognition}
\author{
  Seungyeon Kim$^{1}$ \: Junsu Ha$^{1}$ \: Young Hun Kim$^{1}$ \: Yonghyeon Lee$^{2}$ \: Frank Chongwoo Park$^{1}$\\
  $^{1}$Seoul National University, $^{2}$Massachusetts Institute of Technology\\
  \texttt{\{ksy, hajunsu, yhun\}@robotics.snu.ac.kr, yhl@mit.edu, fcp@snu.ac.kr}
}

\begin{document}
\maketitle

\input{00abstract/camera_ready}

\input{01introduction/camera_ready}

\input{02relatedworks/camera_ready}

\input{03preliminaries/camera_ready}

\input{04screwsplatting/camera_ready}

\input{05experiment/camera_ready}

\input{06conclusion/v1}

\clearpage

\input{00limitation/v1}

\acknowledgments{S. Kim, J. Ha, Y. H. Kim, and F. C. Park were supported
in part by IITP-MSIT under Grant RS-2021-II212068 (SNU AI Innovation Hub),
in part by IITP-MSIT under Grant 2022-220480 and Grant RS-2022-II220480 (Training and Inference Methods for Goal Oriented AI Agents),
in part by IITP-MSIT under Grant RS-2024-00436680 (Collaborative Research Projects with Microsoft Research) through Global Research Support Program in the Digital Field program, 
in part by KIAT under Grant P0020536 (HRD Program for Industrial Innovation), 
in part by SRRC NRF under Grant RS-2023-00208052,
in part by SNU-IPAI, 
in part by SNU-AIIS, 
in part by SNU-IAMD, 
in part by SNU Institute for Engineering Research, 
in part by Hyundai Motor Company and Kia, and 
in part by Microsoft Research Asia.
}

\bibliography{references}  

\clearpage
\section*{Appendix}
\appendix
\input{00appendix/appendix}

\end{document}

%% file: 00abstract/camera_ready.tex

\begin{abstract}
Articulated object recognition -- the task of identifying both the geometry and kinematic joints of objects with movable parts -- is essential for enabling robots to interact with everyday objects such as doors and laptops. However, existing approaches often rely on strong assumptions, such as a known number of articulated parts; require additional inputs, such as depth images; or involve complex intermediate steps that can introduce potential errors -- limiting their practicality in real-world settings. In this paper, we introduce {\it ScrewSplat}, a simple end-to-end method that operates solely on RGB observations. Our approach begins by randomly initializing screw axes, which are then iteratively optimized to recover the object’s underlying kinematic structure. By integrating with Gaussian Splatting, we simultaneously reconstruct the 3D geometry and segment the object into rigid, movable parts. We demonstrate that our method achieves state-of-the-art recognition accuracy across a diverse set of articulated objects, and further enables zero-shot, text-guided manipulation using the recovered kinematic model. See the project website at: \url{https://screwsplat.github.io}.
\end{abstract}

\keywords{Articulated objects, Gaussian splatting, Screw theory} 

%% file: 01introduction/camera_ready.tex
\section{Introduction}
\begin{wrapfigure}{r}{0.5\textwidth}
    \centering
    \vspace{-13pt}
    \includegraphics[width=1\linewidth]{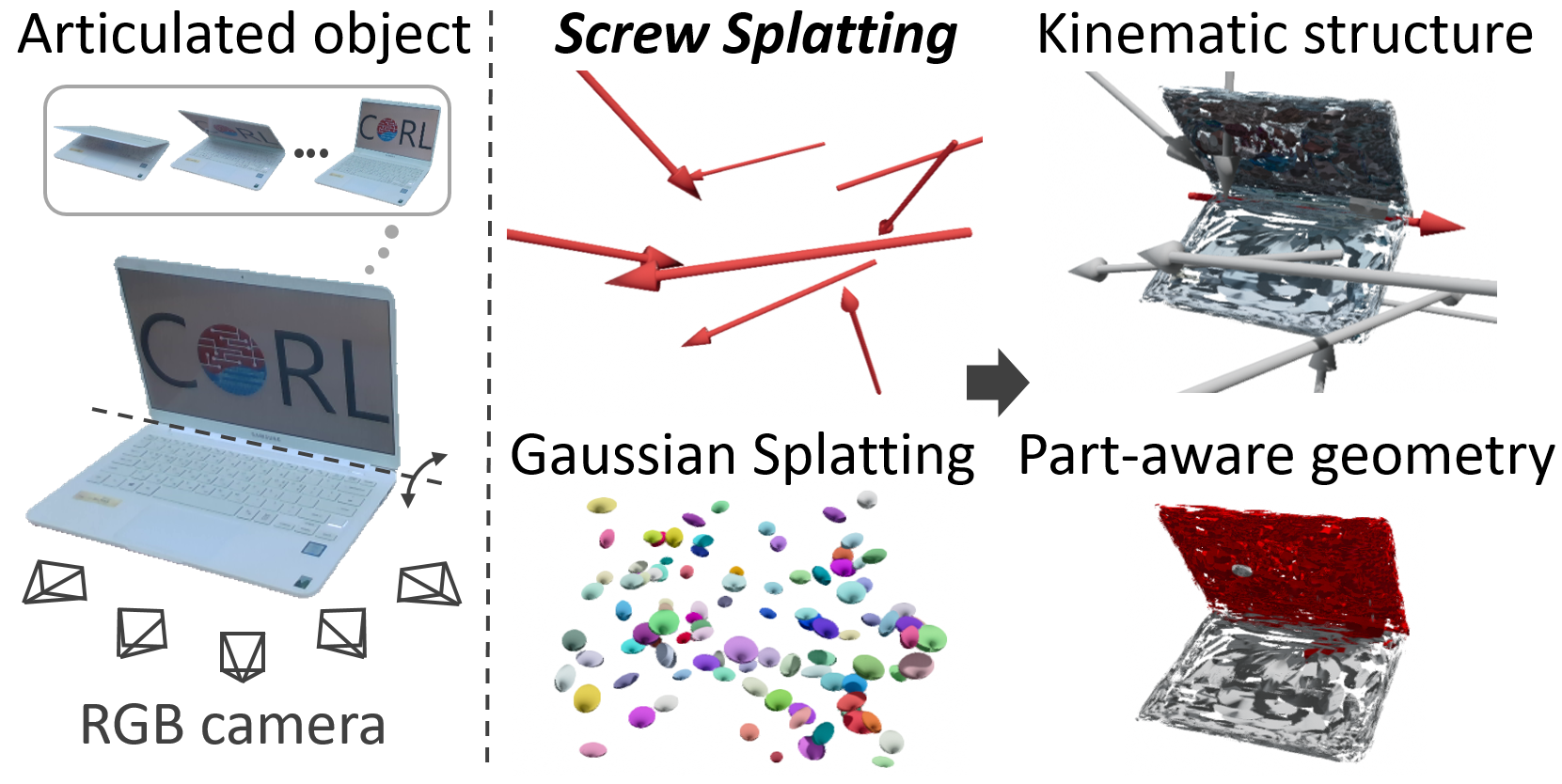}
    \vspace{-13pt}
    \caption{Articulated object recognition by splatting screw axes and Gaussians.}
    \vspace{-12pt}
    \label{fig:intro}
\end{wrapfigure}
Articulated objects with movable parts -- such as doors, laptops, and drawers -- are common in everyday environments, and manipulating them requires understanding both their 3D geometry and underlying kinematic structure (e.g., joint types and axes). While prior work has addressed this using large-scale datasets of 3D objects with annotated joint axes in supervised settings~\cite{li2020category, zeng2021visual, jain2021screwnet, mu2021sdf, tseng2022cla, jain2022distributional, wei2022self, jiang2022ditto, heppert2022category, heppert2023carto, lei2023nap}, such methods struggle to generalize to unseen categories -- a natural limitation of supervised learning. In this work, we tackle a more challenging yet practical scenario: inferring kinematic structure directly from multi-view RGB images under varying object configurations, without relying on category-specific supervision (see the left of Figure~\ref{fig:intro}).

Spurred in part by the success of neural rendering-based 3D reconstruction methods that require no supervised training~\cite{mildenhall2021nerf, wang2021neus, kerbl20233d, wen2023bundlesdf}, recent works have adapted these frameworks for articulated object recognition~\cite{liu2023paris, weng2024neural, kerr2024robot, liu2025building, guo2025articulatedgs}, achieving promising results using raw RGB observations.
However, a key drawback of these methods lies in their reliance on strong assumptions, such as a known number of articulated components or predefined joint types.
Moreover, they often involve multi-stage pipelines with intermediate procedures like point correspondence matching or part clustering, which not only increase overall complexity but also introduce potential points of failure -- errors in these stages can significantly impair final performance.
Furthermore, some of these approaches rely on auxiliary depth inputs, which are often noisier than RGB images -- particularly for transparent or reflective surfaces -- thereby limiting their robustness in real-world scenarios.

We propose \textit{ScrewSplat}, a simple end-to-end method for articulated object recognition that avoids intermediate steps, auxiliary data, and any prior knowledge of joint types or counts. 
We formulate the task as a joint optimization over part-aware geometry, joint axes and types, and joint angles, such that the rendered views match the observations. 
This problem is particularly challenging due to its hybrid structure -- continuous variables (geometry and joint angles) are coupled with discrete, combinatorial ones (part segmentation labels, joint types, and joint counts).

Our key idea is to adopt Gaussian splatting to represent geometry and appearance~\cite{kerbl20233d}, and extend it with a screw model that provides a continuous parameterization of joint axes~\cite{lynch2017modern}.
To represent joint types and counts without relying on discrete variables, we introduce confidence scores over screw axes and use a probability simplex to softly assign Gaussians to rigid parts.
This unified formulation enables smooth, end-to-end optimization over both geometric and kinematic components.
As a result, our method achieves accurate recovery of 3D geometry, screw axes, and part decomposition of articulated objects (see the right of Figure~\ref{fig:intro}).

Building on this framework, we develop a simple yet effective algorithm for articulated object manipulation.
A core strength of our model is its rendering fidelity, which we leverage alongside a large vision-language model to infer target configurations -- such as the joint angles corresponding to an ``open laptop'' or ``folded chair.'' 
These inferred goals are then used to control a robotic manipulator.
Experiments show that \textit{ScrewSplat} consistently outperforms prior methods in recovering both geometric and kinematic structures across a wide range of single- and multi-joint objects. We also demonstrate its effectiveness in zero-shot, text-guided manipulation tasks, validating its practical utility in real-world robotic scenarios.

%% file: 02relatedworks/camera_ready.tex
\section{Related Works}

This section focuses on articulated object recognition methods that {\it do not} rely on explicit supervision~\cite{liu2023paris, weng2024neural, kerr2024robot, liu2025building, guo2025articulatedgs}. A detailed review of supervised methods is provided in Appendix A.1. 

A representative early work is PARIS~\cite{liu2023paris}, which achieves part-level reconstruction and articulation discovery with NeRF, but is restricted to single-joint objects with known joint types. DTA~\cite{weng2024neural}, a more recent approach, extend to multi-joint objects by reconstructing meshes from RGB-D data and inferring kinematic structures via correspondence matching; however, they still depend on depth input and prior knowledge of the number of movable parts. Gaussian splatting-based methods, including ArtGS~\cite{liu2025building} and ArticulatedGS~\cite{guo2025articulatedgs}, provide an alternative. While ArtGS maintains assumptions similar to DTA, ArticulatedGS relaxes them, but remains limited to recovering only one articulation per optimization step. A more detailed review of these approaches is provided in Appendix A.2.

Collectively, while these methods demonstrate strong performance in a category-agnostic setting, they suffer from a critical limitation -- reliance on strong assumptions, such as prior knowledge of the number of articulated joints or even predefined articulation types. This limitation arises from their common strategy of decomposing the problem into the discovery of static and movable object parts, followed by analyzing the motion of the movable parts~\cite{weng2024neural, kerr2024robot, liu2025building, guo2025articulatedgs}. Accurate part discovery generally requires prior knowledge of the number of parts to achieve meaningful segmentation, which restricts the applicability of these methods in real-world scenarios.

%% file: 03preliminaries/camera_ready.tex
\section{Preliminaries}
In this section, we introduce two core components of our approach.
First, we briefly review screw theory -- a fundamental concept in robotics and rigid-body kinematics that enables efficient modeling of joint motion and spatial transformations.
Next, we outline 3D Gaussian splatting, a recent technique for representing and rendering 3D scenes with anisotropic Gaussian primitives.

\subsection{Screw Theory}
Screw theory provides a natural mathematical formulation for describing the motion of a screw, which involves rotation about an axis combined with translation along the axis~\cite{lynch2017modern}. For a given reference frame, a screw axis $\mathcal{S}$ is written as a six-dimensional vector given by:
\begin{equation}
\mathcal{S} = \begin{bmatrix} \omega \\ v \end{bmatrix} \in \mathbb{R}^6,
\end{equation}
where $\omega \in \mathbb{R}^3$ and $v \in \mathbb{R}^3$, and they satisfy either (i) $||\omega|| = 1$ or (ii) $\omega = 0$ and $||v|| = 1$. 

\begin{wrapfigure}{r}{0.5\textwidth}
    \centering
    \vspace{-14pt}
    \includegraphics[width=1\linewidth]{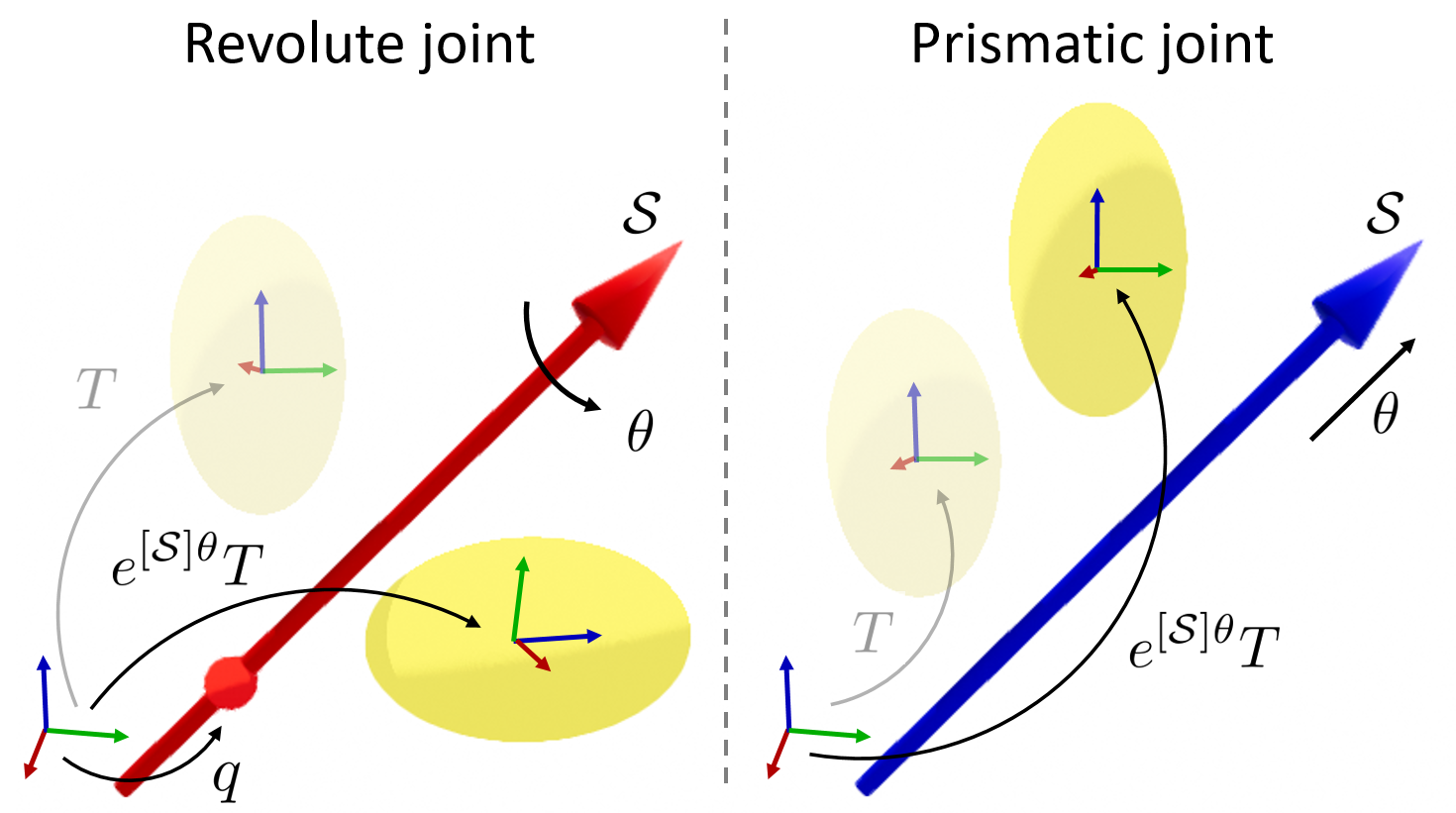}
    \caption{Revolute and prismatic screws axes.}
    \vspace{-15pt}
    \label{fig:screws}
\end{wrapfigure}
If (i) holds, the screw represents a motion consisting of rotation about the axis $\omega$ combined with translation along the same axis. In particular, it satisfies $v = -\omega \times q + h\omega$,
where $q$ is an arbitrary point on the screw axis and $h$ is the pitch of the screw. In this paper, we set $h = 0$, as we consider only {\it revolute joints}, which represent pure rotational motion without pitch. If (ii) holds, the screw represents a motion of pure translation along the axis $v$, and in this case, it corresponds to a {\it prismatic joint}. 

Given a screw axis $\mathcal{S}$ and a joint angle $\theta$, the motion of an arbitrary rigid body coordinate $T \in \mathrm{SE}(3)$ along the screw axis can be expressed using the matrix exponential:
\begin{equation}
T' = e^{[\mathcal{S}]\theta}T,
\end{equation}
where $T' \in \mathrm{SE}(3)$ denotes the transformed rigid body coordinate, and $[\mathcal{S}]$ is the $4 \times 4$ matrix representation of the screw axis $\mathcal{S}$, defined as
\begin{equation}
[\mathcal{S}] = \begin{bmatrix} [\omega] & v \\ 0 & 0\end{bmatrix} \in \mathbb{R}^{4 \times 4}, 
\:\:\:
[\omega] = \begin{bmatrix} 0 & -\omega_3 & \omega_2\\ \omega_3 & 0 & -\omega_1 \\ -\omega_2 & \omega_1 & 0 \end{bmatrix} \in \mathbb{R}^{3 \times 3},
\end{equation}
where $\omega = (\omega_1, \omega_2, \omega_3)$. Figure~\ref{fig:screws} illustrates the revolute and prismatic screw axes and their corresponding screw motions. The matrix exponential formulation admits closed-form expressions for both revolute and prismatic screw axes; further details are provided in Appendix B.1.

\subsection{3D Gaussian Splatting}
\label{sec:3dgs}
3D Gaussian splatting was developed for novel-view synthesis from multiple RGB images and can also be used to obtain a 3D representation of scenes~\cite{kerbl20233d}. It represents a scene using a set of 3D Gaussians, where the $i$th Gaussian $\mathcal{G}_i$ is parameterized by the tuple $(T_i, s_i, \sigma_i, c_i)$. Here, $T_i = [R_i, {\bf \mu}_i] \in \mathrm{SE}(3)$ denotes the pose of the Gaussian, comprising its position $\mu_i \in \mathbb{R}^3$ and orientation $R_i \in \mathrm{SO}(3)$. The parameter $s_i \in \mathbb{R}_+^{3}$ specifies the scale, with the covariance matrix of the Gaussian given by $\Sigma_i = R_i \mathrm{diag}(s_i)^2 R_i^T$. The term $\sigma_i \in [0, 1]$ denotes the opacity, while $c_i$ represents the surface color of the Gaussian ellipsoid, defined by spherical harmonics coefficients. 

To render an RGB image from the colored Gaussians, a typical $\alpha$-blending approach is used, where $\alpha_i$ is a scaled Gaussian function of the $i$th Gaussian $\mathcal{G}_i$, defined in 3D space $\mathbb{R}^3$ as
\begin{equation}
\alpha_i({\bf x}) = \sigma_i e^{-\frac{1}{2}({\bf x}-\mu_i)^T \Sigma_i^{-1} ({\bf x}-\mu_i)},
\end{equation}
where ${\bf x} \in \mathbb{R}^3$. The final color $C$ of a pixel is then computed by blending $\mathcal{N}$ ordered Gaussians overlapping the pixel: \begin{equation}
C = \sum_{i \in \mathcal{N}} c_i \alpha_i \prod_{j=1}^{i-1}(1-\alpha_j).
\end{equation}
The Gaussians are then optimized to minimize the rendering loss function $\mathcal{L}_{\text{render}} = (1 - \lambda) \mathcal{L}_1+ \lambda \mathcal{L}_{\text{D-SSIM}}$, where $\mathcal{L}_1$ and $\mathcal{L}_{\text{D-SSIM}}$ denote the L1 loss and the D-SSIM loss between the rendered RGB image and the ground-truth RGB image, respectively. The weight $\lambda$ is typically set to 0.2.


%% file: 04screwsplatting/camera_ready.tex
\section{ScrewSplat: Integrating Screw Model with 3D Gaussians}
This section introduces \textit{ScrewSplat}, a smooth and differentiable formulation of the rendering-based joint optimization problem over part-aware geometry, joint axes, and joint types of articulated objects.
We first outline the detailed formulation of ScrewSplat along with the associated optimization procedure.
In the following subsection, we present a simple yet effective algorithm for articulated object manipulation that leverages the optimized ScrewSplat as an RGB-based renderer.

\subsection{ScrewSplat}

\begin{figure}[!t]
    \centering
    \includegraphics[width=\linewidth]{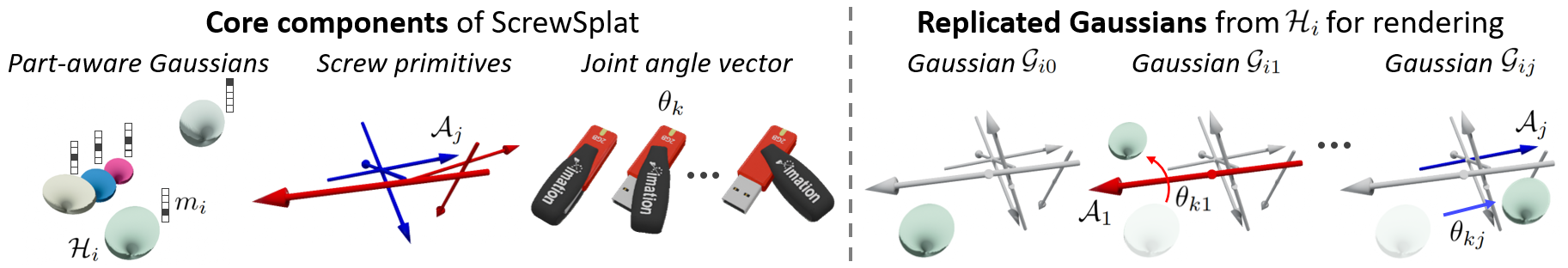}
    \vspace{-15pt}
    \caption{Core components of ScrewSplat ({\it Left}) and the replicated Gaussians derived from the part-aware Gaussian primitive $\mathcal{H}_i$ ({\it Right}).
    }
    \label{fig:screwsplat}
\end{figure}

In this section, we describe ScrewSplat, including (i) its core components, (ii) the RGB rendering procedure, (iii) the loss function used for optimization, and (iv) additional implementation details.
To avoid ambiguity in notation, we first establish our indexing conventions: let $i$ denote the index for Gaussians, $j$ the index for screw axes, and $k$ the index for joint angles of the articulated object. We assume that the observations consist of multi-view RGB images captured under $n_a$ object configurations, such that the index $k$ ranges from from 1 to $n_a$. Additionally, we assume that all movable parts articulate with respect to a single {\it static} base part; that is, we do not consider chain structures.

{\bf Core Components of ScrewSplat.} 
First, we define $n_s$ {\it screw primitives}, where the $j$th screw primitive $\mathcal{A}_j$ is parametrized by a tuple $(\mathcal{S}_j, \gamma_j)$, with $\mathcal{S}_j \in \mathbb{R}^6$ representing a screw axis and $\gamma_j \in [0, 1]$ denoting the confidence. Next, we define $n_g$ {\it part-aware Gaussian primitives}, where the $i$th primitive $\mathcal{H}_i$ is parametrized by an augmented tuple $(T_i, s_i, \sigma_i, c_i, m_i)$. Here, $m_i = (m_{i0}, \cdots, m_{in_s}) \in \Delta^{n_s}$ is a probability simplex over $(n_s + 1)$ parts. Specifically, $m_{i0}$ denotes the probability that the Gaussian belongs to the static base part, and $m_{ij}$ for $j \geq 1$ denotes the probability that the Gaussian is associated with the part whose motion is dominated by the $j$th screw primitive $\mathcal{A}_j$. Lastly, we assign the {\it joint angle vector} $\theta_k = (\theta_{k1}, \cdots, \theta_{kn_s})\in \mathbb{R}^{n_s}$ for RGB observations under $k$'th configuration of the articulated object. The overall core components are illustrated in the left of Figure~\ref{fig:screwsplat}.

{\bf RGB Rendering Procedure with ScrewSplat.}
The key idea behind the RGB rendering procedure is to {\it replicate} Gaussians from each part-aware Gaussian primitive and assign each replicated Gaussian to either the static base or one of the movable parts. Specifically, we replicate $(n_s + 1)$ Gaussians $\mathcal{G}_{ij}$ from the $i$th part-aware Gaussian primitive $\mathcal{H}_i$, where $j = 0, \cdots, n_s$. Each Gaussian $\mathcal{G}_{ij}$ is assigned to the base part if $j = 0$, and to a movable part associated with screw primitive $\mathcal{A}_j$ if $j \geq 1$. Given a joint angle vector $\theta_k$, the replicated Gaussians are parameterized as:
\begin{equation}
    \mathcal{G}_{i0} = (T_i, s_i, \sigma_im_{i0}, c_i), 
    \:\:\:
    \mathcal{G}_{ij} = (e^{[\mathcal{S}_j]\theta_{kj}}T_i, s_i, \sigma_i\gamma_jm_{ij}, c_i),
    \:\:
    \text{for }1 \leq j \leq n_s . 
\end{equation}
The scale and surface color of each replicated Gaussian are inherited from $s_i$ and $c_i$ of $\mathcal{H}_i$, respectively. The pose is $T_i$ for the base part and $e^{[\mathcal{S}_j]\theta_{kj}}T_i$ for movable parts associated with screw primitive $\mathcal{A}_j$. The opacity is derived by scaling the base opacity $\sigma_i$ with the part probability $m_{ij}$, and further modulated by the screw confidence $\gamma_j$ for movable parts. After replicating a total of $n_g \cdot (n_s + 1)$ Gaussians from the $n_g$ part-aware Gaussians, the final RGB image is rendered using an $\alpha$-blending approach. An illustration of the replicated Gaussians is shown on the right of Figure~\ref{fig:screwsplat}.

{\bf Loss Function.} The part-aware Gaussian primitives, screw primitives, and joint angles are jointly optimized to minimize the following loss function:
\begin{equation}
\mathcal{L} = \mathcal{L}_{\text{render}} + \beta \sum_{j} \sqrt{\gamma_j},
\end{equation}
where $\mathcal{L}_{\text{render}}$ is the rendering loss described in Section~\ref{sec:3dgs}, and $\beta$ is a weighting coefficient set to 0.002. The second term serves as a regularization term -- referred to as the {\it parsimony loss} -- which encourages ScrewSplat to represent articulated objects using the smallest possible number of screw primitives.
This term not only pushes the model to select a minimal set of screws, but also promotes the identification of the most reliable ones.

{\bf Implementation Details.} For the part-aware Gaussians, we adopt the same initialization scheme as in Gaussian Splatting, with the exception that the part probabilities $m_i \in \Delta^{n_s}$ are initialized as a uniform distribution.
For the screw primitives, the screw axes $\mathcal{S}_j$ are randomly initialized, and all screw confidences $\gamma_j$ are initialized to 0.9. All joint angle vectors $\theta_k$ are initialized as zero vectors. 

The optimization procedure generally follows that of the original Gaussian Splatting, with a few additional modifications.
The most important modification is that all screw confidences $\gamma_j$ and part probabilities $m_i$ are periodically reset to 0.9 and uniform distribution, respectively. This periodic reinitialization helps ScrewSplat effectively discover meaningful screw primitives.
After optimization, only the screw primitives that satisfy certain criteria remain, ensuring that only the most relevant ones are retained. 
Further optimization details can be found in Appendix B.2.

\subsection{Controlling Joint Angles Using ScrewSplat as a Renderer}

\begin{figure}[!t]
    \centering
    \includegraphics[width=\linewidth]{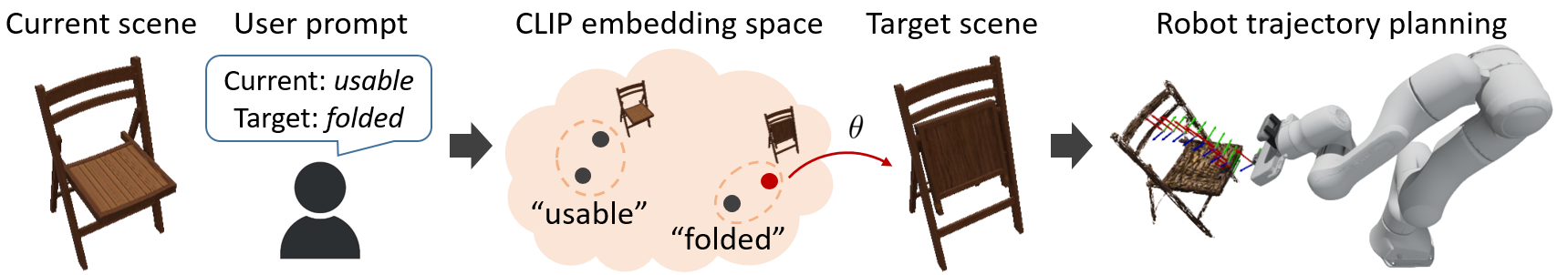}
    \vspace{-15pt}
    \caption{The problem setting for text-guided articulated object manipulation ({\it Left}), optimization of the target joint angles using CLIP ({\it Middle}), and corresponding robot trajectory planning ({\it Right}).
    }
    \vspace{-10pt}
    \label{fig:manipulation}
\end{figure}

The optimized ScrewSplat serves as an RGB image renderer conditioned on the joint angle vector $\theta$; that is, the visual appearance (i.e., RGB image) of the articulated object $I$ from an arbitrary camera pose can be obtained through a continuous -- and even differentiable -- function $\pi$, such that $I = \pi(\theta)$. This function $\pi$ enables a variety of applications, such as estimating the current pose of the articulated object, by defining an appropriate objective function on the rendered image and optimizing the joint angles accordingly~\cite{liu2024differentiable}.

In this paper, we primarily focus on controlling the joint angles of an articulated object to match a given text prompt using visual foundation models. Specifically, given the current visual appearance of the object $I_c$ and a text description $t_c$ of its current state, along with a target text prompt $t_p$, our goal is to find a joint angle vector $\theta$ such that the rendered appearance $I = \pi(\theta)$ aligns with the target prompt $t_p$. The problem setting is illustrated on the left of Figure~\ref{fig:manipulation}.

To achieve this, we utilize the CLIP model, which embeds both RGB images and text into a shared latent space, enabling the computation of similarity between visual and textual inputs~\cite{radford2021learning}. Let $e_t$ and $e_I$ denote the pretrained text and image encoders of CLIP, respectively. In this paper, we adopt a directional CLIP loss~\cite{gal2022stylegan}, defined as:
\begin{equation}
    \mathcal{L}_{\text{CLIP-dir}} = 1 - \frac{\triangle I (\theta) \cdot \triangle T}{||\triangle I (\theta)|| \: ||\triangle T||},
\end{equation}
where $\triangle I (\theta) = e_I(\pi(\theta)) - e_I(I_c)$ and $\triangle T = e_T(t_p) - e_T(t_c)$ represent the directional shifts in the CLIP latent space. We optimize joint angle vector $\theta$ using Bayesian optimization. This process is illustrated in the middle of Figure~\ref{fig:manipulation}.

After obtaining the target joint angle vector $\theta$, we plan a simple trajectory for the robot end-effector tip and generate a kinematically feasible robot trajectory accordingly, as shown on the right of Figure~\ref{fig:manipulation}.
Further details of the optimization and planning procedure are provided in Appendix B.3.

%% file: 05experiment/camera_ready.tex
\section{Experiments}

In this section, we empirically demonstrate that (i) ScrewSplat outperforms existing state-of-the-art methods in recognizing articulated objects, and (ii) our method can be effectively applied to text-guided articulated object manipulation in both simulated and real-world settings, followed by successful robot-object interaction.

\begin{figure}[]
    \centering
    \includegraphics[width=\linewidth]{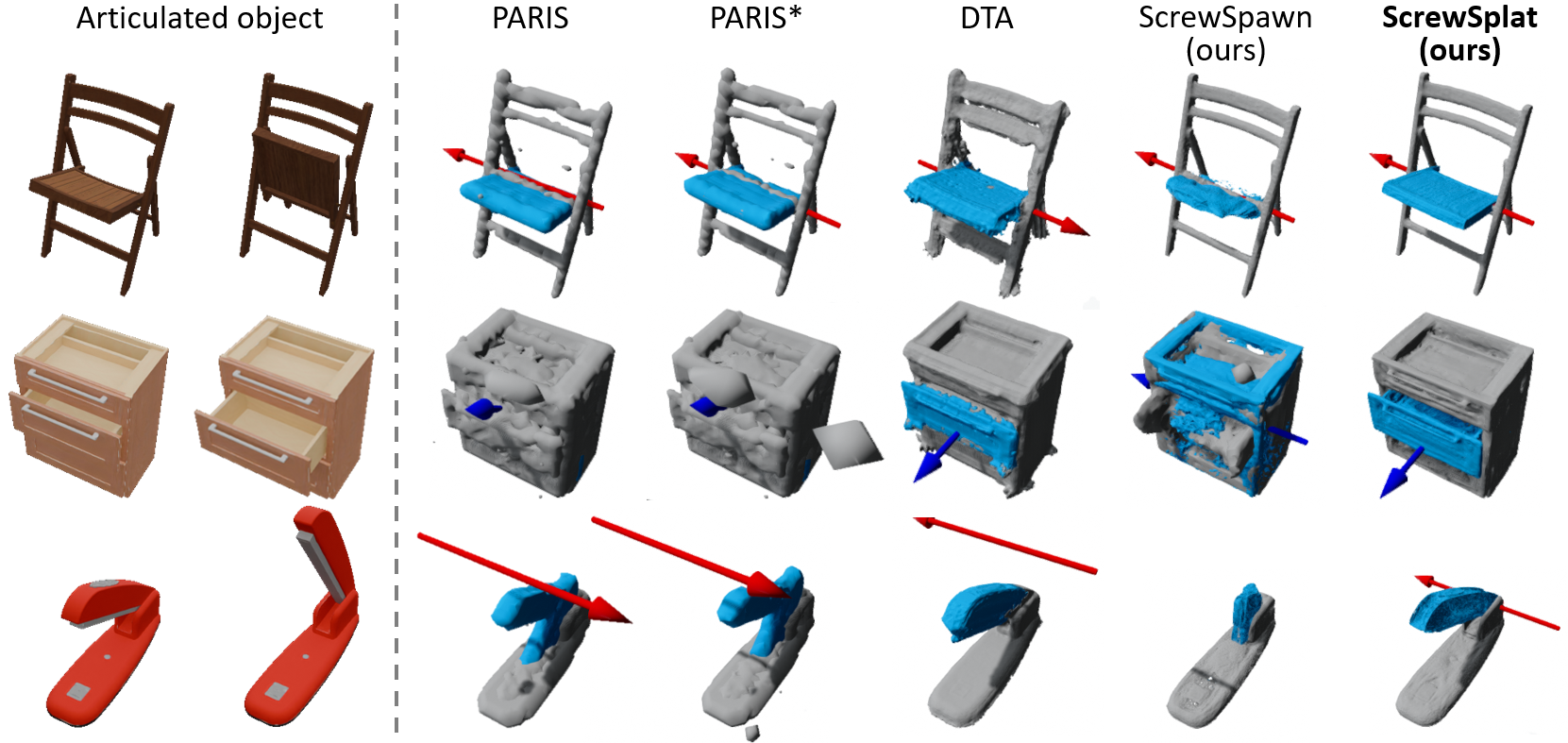}
    \vspace{-15pt}
    \caption{Reconstructed meshes and screw axes for single-joint objects using each method. Static parts are shown in gray, movable parts in cyan, revolute joints in red, and prismatic joints in blue.}
    \vspace{-5pt}
    \label{fig:single_screw_vis}
\end{figure}

\begin{table}[]
\centering
\caption{Recognition performance for single-joint objects, averaged across all instances.
Object-wise recognition performance is provided in Appendix D.1.}
\label{table:single_screw_eval}
\begin{tabular}{lccccccccccccc}
\multicolumn{1}{c}{} &\multicolumn{3}{c}{Geometry ($\downarrow$)} &\multicolumn{1}{c}{} &\multicolumn{2}{c}{Motion ($\downarrow$)} &\multicolumn{1}{c}{} &\multicolumn{2}{c}{Appearance ($\uparrow$)} \\
\cline{2-4}
\cline{6-7} 
\cline{9-10} 

\multicolumn{1}{c}{\bf METHOD}  &\multicolumn{1}{c}{CD-s} &\multicolumn{1}{c}{CD-m} &\multicolumn{1}{c}{CD-w} &\multicolumn{1}{c}{} &\multicolumn{1}{c}{Ang.} &\multicolumn{1}{c}{Pos.} &\multicolumn{1}{c}{} &\multicolumn{1}{c}{PSNR} &\multicolumn{1}{c}{SSIM} \\

\hline 
PARIS~\cite{liu2023paris}        & 54.015 & 18.032 & 40.192 &   & 17.656 & 2.020 &  & 28.64 & 0.970\\
PARIS*~\cite{liu2023paris}        & 49.706 & 8.864 & 33.856 &   & 16.287 & 1.742 &  & 28.66 & 0.970\\
DTA~\cite{weng2024neural}       & 0.538 & 0.528 & 0.360 &  &  0.437 & 0.308 &  & - & -\\
ScrewSpawn (ours) & 0.617 & 11.566 & 0.946 &   & 24.869 & 0.902 &  & 29.11 & 0.982 \\
ScrewSplat (ours)  & \textbf{0.319} & \textbf{0.211} & \textbf{0.261} &  & \textbf{0.084} & \textbf{0.010} & & \textbf{38.07} & \textbf{0.993} \\
\hline
\end{tabular}
\end{table}

{\bf Baseline Methods.} 
We compare ScrewSplat with the following baselines: {\it PARIS}~\cite{liu2023paris}, {\it PARIS*}, {\it DTA}~\cite{weng2024neural}, and {\it ScrewSpawn}. PARIS and PARIS* (PARIS augmented with depth data) recover a single joint axis assuming a known joint type, while DTA discovers a predefined number of joint axes.
Since only DTA supports multiple joint axes, we compare against DTA for multi-joint objects.
These methods recover kinematic structures from multi-view images under two different object configurations; PARIS uses RGB input, while PARIS* and DTA use RGB-D.
We also introduce ScrewSpawn, an ablation model that follows the ScrewSplat framework but spawns only a single screw (with a known joint type).
This model is used to validate the necessity of ``splatting'' multiple screw primitives in ScrewSplat.
Detailed implementations can be found in Appendix C.1.

{\bf Dataset.} 
We select ten single-joint objects and three multi-joint objects from distinct categories in the PartNet-Mobility dataset~\cite{xiang2020sapien}.
Using Blender~\cite{blender}, a photorealistic renderer, we obtain multi-view RGB images and depth images (used for training PARIS* and DTA) under varying object configurations.
We place 48 camera poses uniformly over a hemisphere centered on each object.
For ScrewSplat and ScrewSpawn, we use images captured under five configurations, while for PARIS, PARIS*, and DTA, we use images from two configurations selected from the same five.
For single-joint objects, the five configurations correspond to evenly spaced joint angles along the motion range.
For multi-joint objects, five randomly sampled joint angle vectors are used.
Further details for the evaluation dataset are provided in Appendix C.1.

\subsection{Articulated Object Recognition Performance}
We compare the performance of ScrewSplat with the baselines.
To evaluate the recognition quality of articulated objects, we adopt three types of metrics: {\it geometry}, {\it motion}, and {\it appearance}. The geometry metric includes the bi-directional Chamfer-$l2$ distance between point clouds sampled from the reconstructed and ground-truth meshes. We report this metric separately for the static part (CD-s, mm), the movable parts (CD-m, mm), and the entire object (CD-w, mm). The motion metric includes the angular error (Ang., $^\circ$) between the estimated and ground-truth screw axes, and the axis position error (Pos., 0.1m) for revolute joints, calculated as the minimum distance between corresponding screw axes. The appearance metric includes Peak Signal-to-Noise Ratio (PSNR) and Structural Similarity Index Measure (SSIM), computed on rendered images under unseen joint angles.
Details of the evaluation metrics and their computation are available in Appendix C.1.

\begin{wrapfigure}{r}{0.5\textwidth}
    \centering
    \vspace{-15pt}
    \includegraphics[width=1\linewidth]{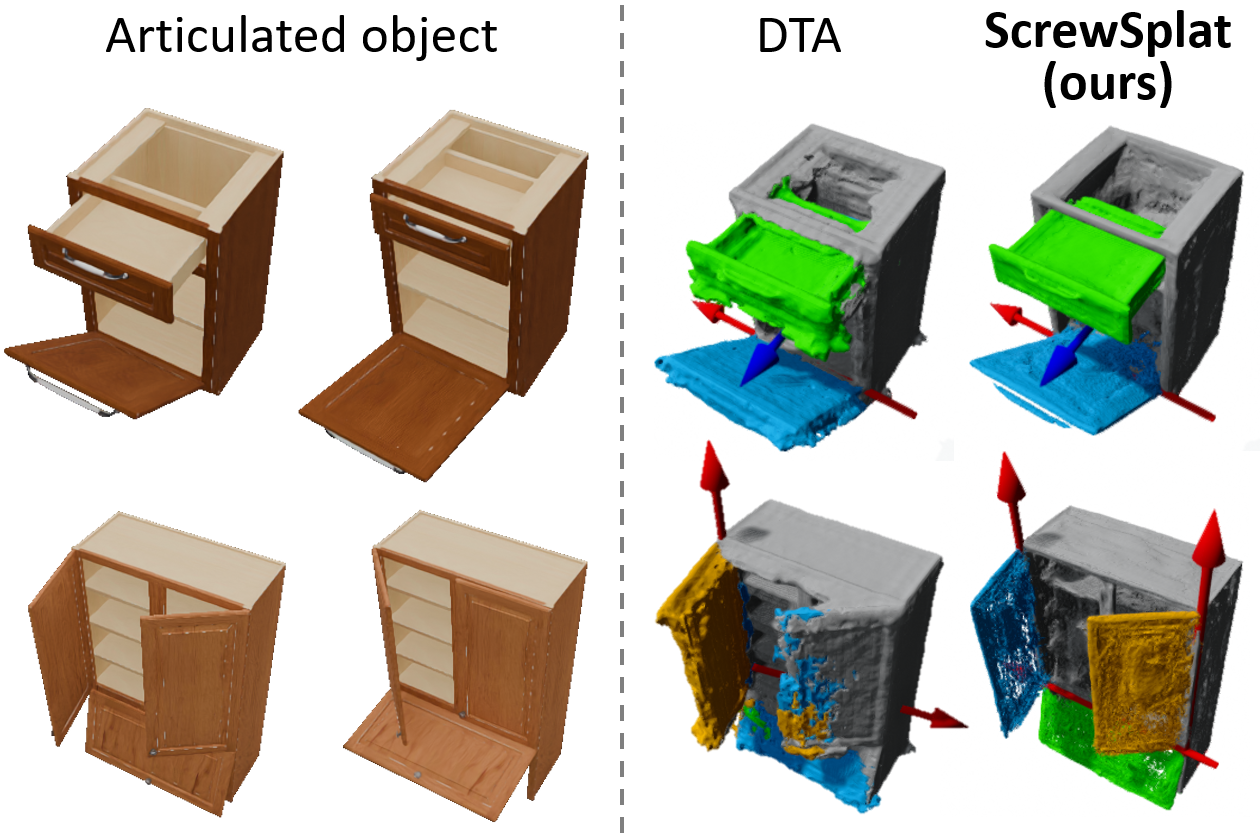}
    \caption{Reconstructed meshes and screw axes for multi-joint objects using each method.}
    \vspace{-15pt}
    \label{fig:recognition_multi_results}
\end{wrapfigure}
\textbf{Single-joint Objects.}
Figure~\ref{fig:single_screw_vis} shows the recognition results, including reconstructed meshes and screw axes, on three representative single-joint objects.
PARIS and PARIS* performed well on the folding chair but failed to recognize the other objects.
DTA generally outperforms PARIS, but still fails in certain cases such as the stapler.
ScrewSplat consistently demonstrates the best performance across all cases, both in terms of geometry reconstruction and kinematic structure estimation.
In contrast, ScrewSpawn fails to accurately recover both geometry and kinematic structure in most cases, highlighting the effectiveness of ScrewSplat’s multi-screw formulation. 
Table~\ref{table:single_screw_eval} shows the quantitative recognition results.
We demonstrate that ScrewSplat achieves the highest overall performance across geometry, motion, and visual appearance.
PARIS and PARIS* struggle to recognize almost all of the other objects.
ScrewSpawn is able to reconstruct the static parts to some extent but fails to accurately estimate the movable parts and screw axes.
DTA performs slightly worse than our method but achieves comparable results overall.
Although our method uses data from a wider range of joint configurations, it is particularly notable that it achieves the best performance while relying solely on RGB inputs and without any prior knowledge of the joints.
Detailed object-wise results are provided in Appendix D.1.

\begin{figure}[!t]
    \centering
    \vspace{-15pt}
    \includegraphics[width=\linewidth]{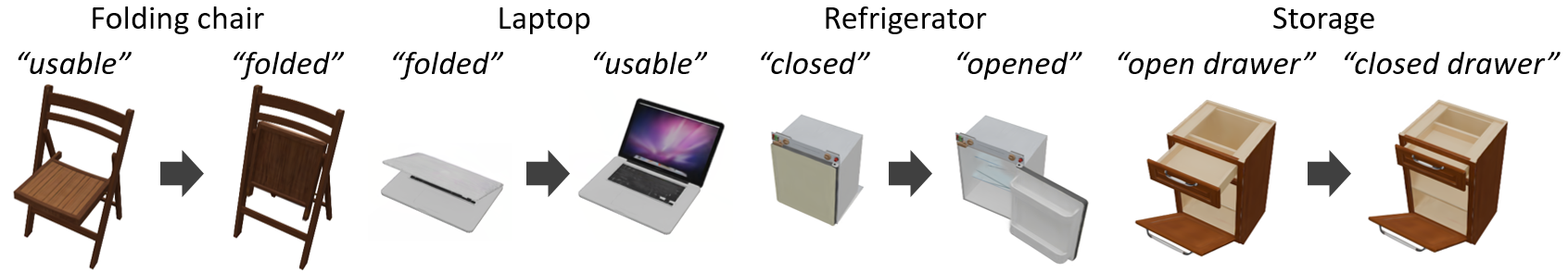}
    \caption{Text-guided articulated object manipulation results in simulation.}
    \vspace{-5pt}
    \label{fig:sim_manipulation_results}
\end{figure}

\begin{table}[]
\centering
\caption{Recognition performance on multi-joint objects, averaged over all instances, geometries, and screw axes.
Detailed object-wise recognition results are provided in Appendix D.1.
}
\label{table:multi_screw_eval}
\begin{tabular}{lcccccccccc}
\multicolumn{1}{c}{} &\multicolumn{3}{c}{Geometry ($\downarrow$)} &\multicolumn{1}{c}{} &\multicolumn{2}{c}{Motion ($\downarrow$)} &\multicolumn{1}{c}{} &\multicolumn{2}{c}{Appearance ($\uparrow$)} \\
\cline{2-4}
\cline{6-7} 
\cline{9-10} 

\multicolumn{1}{c}{\bf METHOD}  &\multicolumn{1}{c}{CD-s} &\multicolumn{1}{c}{CD-m$_m$} &\multicolumn{1}{c}{CD-w} &\multicolumn{1}{c}{} &\multicolumn{1}{c}{Ang$_m$.} &\multicolumn{1}{c}{Pos$_m$.} &\multicolumn{1}{c}{} &\multicolumn{1}{c}{PSNR} &\multicolumn{1}{c}{SSIM} \\

\hline 
DTA~\cite{weng2024neural}       & \textbf{0.568} & 5.647 & \textbf{0.476} &  &  7.233 & 28.877 &  & - & -\\
ScrewSplatting(ours)  & 0.675 & \textbf{0.096} & 0.666 &  &  \textbf{0.130} & \textbf{0.002} &  & \textbf{36.76} & \textbf{0.987}\\
\hline
\end{tabular}
\end{table}

{\bf Multi-joint Objects.}
Figure~\ref{fig:recognition_multi_results} presents the recognition results on multi-joint objects.
DTA successfully predicts both screw axes for objects with two screws (top row of Figure~\ref{fig:recognition_multi_results}), but fails to correctly identify more than one axis in objects with three screws (bottom row).
In contrast, ScrewSplat successfully recognizes all screw axes in both cases.
Quantitative recognition results for multi-joint objects are provided in Table~\ref{table:multi_screw_eval}.
While DTA performs slightly better on static parts and whole-object geometry, ScrewSplat achieves substantially higher accuracy in reconstructing the geometry of movable parts and estimating joint axes.
Overall, ScrewSplat consistently outperforms prior methods in recognizing articulated objects, both for single-joint and multi-joint objects.

\subsection{Articulated Object Manipulation Results}
In this section, we demonstrate the effectiveness of ScrewSplat on the task of text-guided articulated object manipulation in both simulated and real-world settings.

\begin{figure}[!t]
    \centering
    \includegraphics[width=\linewidth]{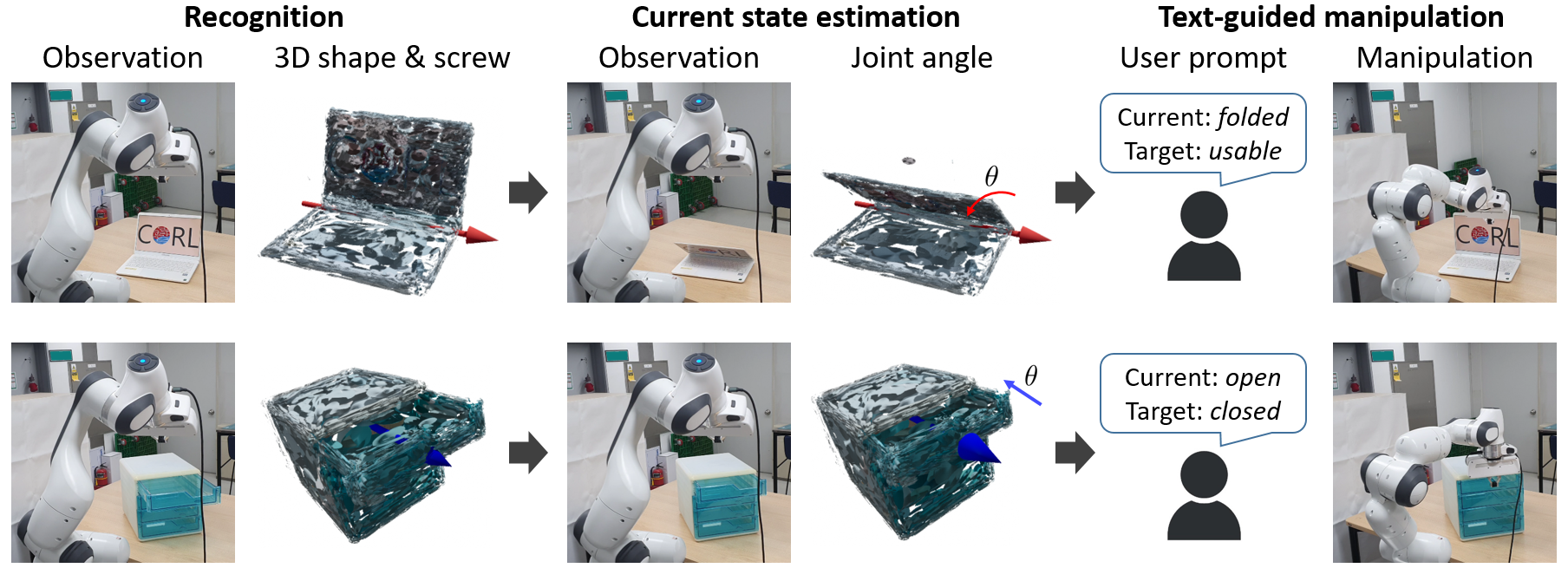}
    \vspace{-12pt}
    \caption{Overall pipeline for text-guided robotic manipulation of the real-world articulated objects.
    }
    \vspace{-10pt}
    \label{fig:real_manipulation_results}
\end{figure}

{\bf Simulated Articulated Object Manipulation.} Figure~\ref{fig:sim_manipulation_results} shows the results of text-guided joint angle optimization using ScrewSplat models optimized on simulated articulated objects.
By providing a suitable description of the current state along with a target text prompt, the objects can be manipulated into the desired configuration.
We observe that for single-joint objects, simple prompts are sufficient to find the desired configurations, whereas multi-joint objects require more specific instructions (e.g., “closed drawer” instead of just “closed”).

{\bf Real-world Articulated Object Manipulation.} 
For real-world scenarios where a robot manipulator physically adjust the joint angles of articulated objects, the overall manipulation pipeline consists of three main stages, as illustrated in Figure~\ref{fig:real_manipulation_results}:
(i) a {\it recognition} step, where ScrewSplat is optimized using multi-view RGB observations collected under several joint configurations manually manipulated by a human;
(ii) a {\it current state estimation} step, which estimates the object’s current joint angle by optimizing an appropriate loss function; and
(iii) a {\it text-guided manipulation} step, which determines the target joint angle based on a given text prompt and executes the corresponding robot manipulation.
As shown on the left of Figure~\ref{fig:real_manipulation_results}, ScrewSplat accurately reconstructs the shape and kinematic structure of real-world objects -- including even a translucent drawer.
A well-trained ScrewSplat further enables precise estimation of the current joint angle and facilitates successful text-guided object manipulation, as shown in the middle and right of Figure~\ref{fig:real_manipulation_results}, respectively.
Further details for real-world text-guided manipulation are provided in Appendix C.2 and D.3, respectively.


%% file: 06conclusion/v1.tex
\section{Conclusion}
We propose ScrewSplat, a novel end-to-end framework for articulated object recognition that operates solely on RGB observations.
By leveraging screw theory and Gaussian splatting, and introducing confidence scores over screw axes along with a part probability simplex for Gaussians, our formulation enables smooth and unified optimization over both geometric and kinematic components.
Unlike prior approaches, ScrewSplat avoids strong assumptions, complex intermediate steps, and reliance on depth data, resulting in a more robust and generalizable solution.
We also demonstrate that ScrewSplat shows state-of-the-art performance in recovering the geometry and kinematic structure of both single- and multi-joint articulated objects.
Furthermore, we show that ScrewSplat can be directly applied to zero-shot, text-guided manipulation of articulated objects, enabling robots to physically adjust joint angles according to high-level user intent in real-world environments.


%% file: 00limitation/v1.tex
\section*{Limitations and Future Directions}

\begin{figure}
    \centering
    \includegraphics[width=\linewidth]{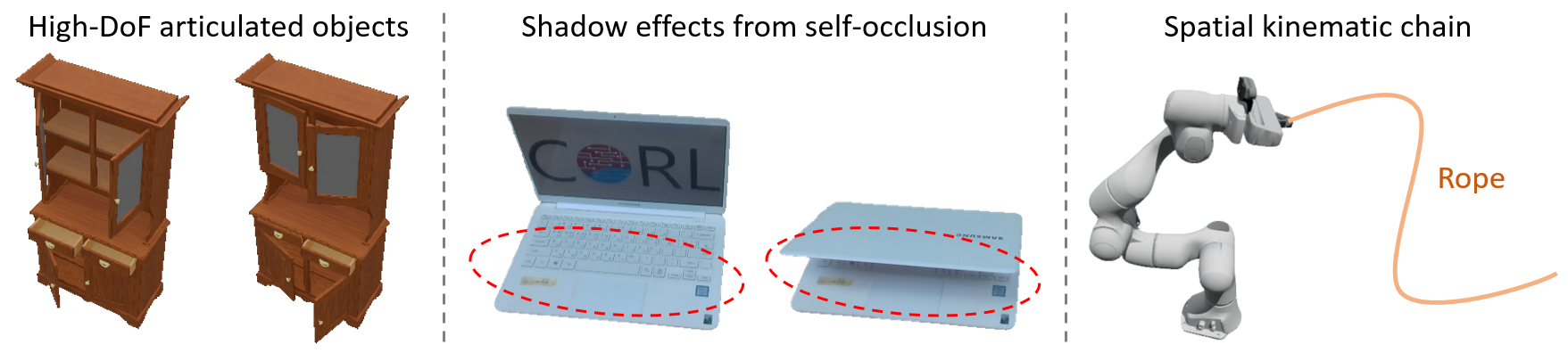}
    \vspace{-12pt}
    \caption{Limitations and Future Directions of ScrewSplat.
    }
    \vspace{-5pt}
    \label{fig:future_works}
\end{figure}

The recognition performance of ScrewSplat is sensitive to the weight $\beta$ of the parsimony loss, particularly for multi-joint objects, making it crucial to find an appropriate $\beta$. 
Specifically, when $\beta$ is set too low, ScrewSplat attempts to recognize the articulated object using many screws, leading to a trivial solution where the same movable part in each configuration is assigned to different screws. 
Conversely, when $\beta$ is set too high, the algorithm fails to detect any movable parts and recognizes the object as static.
Due to these tendencies, ScrewSplat still struggles to identify all reliable joint axes for {\it high-DoF articulated objects}, such as object with six joints shown in the left of Figure~\ref{fig:future_works}. We aim to address this limitation in the future by developing stable optimization techniques. 
A key future direction will be to explore methods for properly initializing the screw axes at the beginning, and to develop adaptive techniques for adjusting $\beta$ based on the current recognition performance during optimization. Additionally, inspired by the optimization techniques in Gaussian Splatting~\cite{kerbl20233d}, we will explore techniques to either remove or densify screw axes during optimization.

One of the challenges we observed in articulated object recognition is accounting for the {\it shadow effects caused by movable parts}. 
As shown in the middle of Figure~\ref{fig:future_works}, different configurations of a movable part cast varying shadow effects on the static parts in the RGB image. In other words, the same part (or the same part-aware Gaussian primitive) should exhibit different colors depending on the joint angle. 
With a sufficiently low $\beta$, when these shadow effects are significant, ScrewSplat may discover additional Gaussians or even identify new parts to model the effect. 
To find the appropriate kinematic structure, it is crucial to model these effects, and this remains an area for future work. 
Possible future directions include formulating a method that explicitly optimizes for light information to model these shadow effects, as suggested in~\cite{bolanos2024gaussian}, and modeling the color of Gaussians as a function of joint configurations -- using additional deformation functions such as Implicit Linear Blend Skinning (LBS)~\cite{liu2024differentiable, loper2023smpl} -- and optimizing it directly. 

Lastly, a promising extension of ScrewSplat is the ability to recognize articulated objects using a {\it spatial kinematic chain}. Consider a kinematic chain in which a screw $\mathcal{S}_1$ is attached to a static base part, and subsequent screws $\mathcal{S}_1$, $\mathcal{S}_2$, ..., $\mathcal{S}_n$ are serially connected. Given the joint angles $\theta_1$, $\theta_2$, ..., $\theta_n$ corresponding to each screw, the motion of an arbitrary rigid body coordinate $T \in \text{SE}(3)$ attached to the $n$th movable part can be described by the following {\it product of exponentials formula}~\cite{lynch2017modern}:
\begin{equation}
T' = e^{[\mathcal{S}_1]\theta_1}e^{[\mathcal{S}_2]\theta_2} \cdots e^{[\mathcal{S}_n]\theta_n}T,
\end{equation}
where $T' \in \text{SE}(3)$ represents the transformed coordinate. 
We have focused on objects where all movable parts articulate with respect to a single static base part, but there are many objects, such as a robotic manipulator as shown in the right of Figure~\ref{fig:future_works}, where modeling a spatial kinematic chain is inevitable. In future work, we aim to extend ScrewSplat by leveraging the product of exponentials formula to recognize the geometry and kinematic structure of spatial chains, such as robots, using only RGB images -- this direction aligns with a recent work that uses full point clouds~\cite{lin2024autourdf}. Furthermore, we plan to explore the recognition of one-dimensional deformable objects such as ropes, in the spirit of pseudo-rigid-body theory, and use the recognized model to develop effective manipulation strategies~\cite{viswanath2023handloom, chi2024iterative}, which will also be a direction for future work.

%% file: 00appendix/appendix.tex


\input{00appendix/A1relatedworks}

\input{00appendix/A2implementation}

\input{00appendix/A3experiment}

\input{00appendix/A4additional}

%% file: 00appendix/A1relatedworks.tex

\section{Extended Related Works}

\subsection{Articulated Object Recognition Using Supervised Learning}

There has been considerable interest in recognizing articulated objects within a supervised learning framework, where deep neural networks are trained to predict articulation parameters (e.g., joint axes and joint angles) directly from raw visual input~\cite{li2020category, zeng2021visual, jain2021screwnet, mu2021sdf, tseng2022cla, jain2022distributional, wei2022self, jiang2022ditto, heppert2022category, heppert2023carto, lei2023nap}. Some of these methods also aim to simultaneously reconstruct the part-aware geometry of objects~\cite{tseng2022cla, jiang2022ditto, heppert2023carto}, which aligns with our objectives. While the problem of 3D recognition of rigid objects is well studied and has numerous established solutions~\cite{varley2017shape, kato2018neural, liu2019learning, park2019deepsdf, van2020learning, ichnowski2021dex, kerr2022evo, kim2022dsqnet, dai2022graspnerf, kim2023leveraging, kim2024t, kim2025dreamgrasp}, the recognition of articulated objects is significantly more challenging, as it requires inferring not only object geometry but also the underlying kinematic structure. Consequently, it remains an open problem.

Early studies often addressed articulated object recognition problem by employing category-driven approaches, using category information to assist in identifying the kinematic structure, and testing on unseen objects belonging to known categories~\cite{li2020category, mu2021sdf, wei2022self, abbatematteo2019learning}. More recently, to handle a broader range of objects beyond those in known categories, category-agnostic recognition approaches have been proposed. These methods primarily aim to reconstruct part-level geometry and kinematic structures, which closely align with our objective. For example, CARTO~\cite{heppert2023carto} predicts the geometry as a defomable signed distance function (SDF) representations -- similar to the category-specific A-SDF~\cite{mu2021sdf} but removes the reliance on object categories. Ditto~\cite{jiang2022ditto} predicts articulable neural occupancy fields to generate digital twins of articulated objects from point cloud observations under two different articulation states. Articulate-Anything~\cite{le2024articulate}, a recent approach, integrates a vision-language foundation model to recognize part-aware geometries and kinematic structures, and incorporates interactable digital twins into simulators for sim-to-real robot learning. They demonstrate generalizability to unseen objects within similar categories, and even to unseen categories through foundation models. However, they inherently struggle to generalize to objects that differ significantly from the training categories.

\subsection{Articulated Object Recognition Using Per-object Optimization Methods}
Several works have attempted to recognize articulated objects by directly fitting 3D representations and kinematic structures to observations without any supervision. Since these methods typically perform optimization for each object individually, they are often referred to as per-object optimization methods. A representative early work in per-object optimization methods is PARIS~\cite{liu2023paris}, which presents a method based on neural radiance fields (NeRF). Specifically, PARIS defines separate radiance fields for movable and static parts, performing joint rendering and optimization to achieve part-level reconstruction and articulation discovery. However, this approach is only applicable to single-joint objects, i.e., articulated objects with only one movable part, with known joint types.

Recently, approaches capable of covering multi-joint objects have been proposed. A notable example is DTA~\cite{weng2024neural}, which first reconstructs two entire meshes using RGB-D data from observations under two configurations of articulated objects, and then determines the kinematic structure using a feature correspondence matching module. During the mesh reconstruction process, depth images are used, and in the feature correspondence matching step, the number of movable parts must also be known. DTA successfully infers the kinematic structure of articulated objects with multiple movable parts using this additional information. Subsequently, research utilizing Gaussian splatting~\cite{kerbl20233d}, such as ArtGS~\cite{liu2025building} and ArticulatedGS~\cite{guo2025articulatedgs}, has emerged as an alternative to neural radiance fields. These approaches are somewhat similar to ours in that they leverage Gaussian splatting. ArtGS, like DTA, utilizes point correspondence matching and similarly requires depth images and knowledge of the number of movable parts. ArticulatedGS, however, does not rely on these assumptions but is still limited to discovering only one articulation information per optimization step.

While these methods demonstrate strong performance without supervision, the review above highlights several limitations. The most important limitation is that they rely on assumptions such as the articulated object has a single joint axis, or the user should know the number of articulated joints, or even predefined articulation types. Some works also rely on auxiliary depth inputs, which are often noisier than RGB images -- particularly for transparent or reflective surfaces -- thus limiting their robustness in real-world scenarios. Finally, it is important to emphasize that these methods often involve multi-stage pipelines with intermediate procedures like point correspondence matching, which not only increase overall complexity, but also contrasts with our simple, end-to-end framework that operates soly on RGB observations and does not rely on such assumptions.

\subsection{Articulation Discovery via Robot-Object Interaction}
Beyond observation-based recognition approaches, several works have explored active interaction strategies that allow a robot to interact with unknown articulated objects and collect additional observations for articulation reasoning~\cite{gadre2021act, hsu2023ditto, ma2023sim2real}. These methods often rely on pre-trained networks to guide interaction, rather than some learning-free strategies that cannot extract joint parameters from static observations. For instance, \cite{gadre2021act} uses an RGB image as input to predict hold and push locations via a trained network. The predicted actions are then executed by a human to modify the articulated object's configuration, and subsequent observations are used to infer the kinematic structure. Similar methods such as \cite{hsu2023ditto} and \cite{ma2023sim2real} also leverage 3D point cloud inputs to predict interaction positions and directions. Using point cloud observation data as 3D geometric information enables the robot to interact with the object and actively acquire additional observations. 

The observation-based recognition approaches described above have drawbacks compared to the interaction-based methods discussed here. One major limitation is that collecting such data typically requires manual manipulation, which is labor-intensive and difficult to automate. While interaction-based methods are not yet perfect -- often still requiring additional human interaction or failing to produce appropriate interactions for novel articulated objects -- we believe that integrating observation-based recognition with interaction-based approaches is a promising future research direction that could lead to more effective recognition methods for articulated objects.


%% file: 00appendix/A2implementation.tex

\newpage
\section{Implementation Details for ScrewSplatting}

\subsection{Details for Screw Theory}
In this section, we describe the closed-form matrix exponential expressions for both revolute and prismatic screw axes~\cite{lynch2017modern}. For convenience, we briefly review the screw theory outlined in Section 3.1. A six-dimensional screw axis $\mathcal{S}$ is given by:
\begin{equation}
\mathcal{S} = \begin{bmatrix} \omega \\ v \end{bmatrix} \in \mathbb{R}^6.
\end{equation}
For a revolute joint, the screw axis satisfies $||\omega|| = 1$ and $v = -\omega \times q$, where $q$ is an arbitrary point on the screw axis. For a prismatic joint, $\omega = 0$ and $||v|| = 1$. Given a screw axis $\mathcal{S}$ and a joint angle $\theta$, the motion of an arbitrary rigid body coordinate $T \in \mathrm{SE}(3)$ along the screw axis can be expressed using the matrix exponential:
\begin{equation}
T' = e^{[\mathcal{S}]\theta}T,
\end{equation}
where $T' \in \mathrm{SE}(3)$ denotes the transformed rigid body coordinate, and $[\mathcal{S}]$ is the $4 \times 4$ matrix representation of the screw axis $\mathcal{S}$, defined as
\begin{equation}
[\mathcal{S}] = \begin{bmatrix} [\omega] & v \\ 0 & 0\end{bmatrix} \in \mathbb{R}^{4 \times 4}, 
\:\:\:
[\omega] = \begin{bmatrix} 0 & -\omega_3 & \omega_2\\ \omega_3 & 0 & -\omega_1 \\ -\omega_2 & \omega_1 & 0 \end{bmatrix} \in \mathbb{R}^{3 \times 3},
\end{equation}
where $\omega = (\omega_1, \omega_2, \omega_3)$. 

In general, the matrix exponential $e^{[\mathcal{S}]\theta}$ is computed using its series expansion:
\begin{eqnarray}
e^{[\mathcal{S}]\theta} &=& I + [\mathcal{S}]\theta + [\mathcal{S}]^2 \frac{\theta^2}{2!} + [\mathcal{S}]^3 \frac{\theta^3}{3!} + \cdots \\ &=&
\begin{bmatrix} e^{[\omega]\theta} & G(\theta)v \\ 0 & 1 
\end{bmatrix},
\end{eqnarray}
where $G(\theta)$ represents the function that generates the translational part of the motion, and is given by:
\begin{equation}
G(\theta) = I\theta + [\omega] \frac{\theta^2}{2!} + [\omega]^2 \frac{\theta^3}{3!} + \cdots
\end{equation}

For a revolute joint, $G(\theta)$ has a closed-form expression using the fact that $[\omega]^3 = [\omega]$:
\begin{equation}
G(\theta) = I\theta + (1-\cos \theta) [\omega] + (\theta - \sin \theta)[\omega]^2.
\end{equation}
Thus, the closed-form matrix exponential expression for the revolute joint is given by:
\begin{equation}
e^{[\mathcal{S}]\theta} = \begin{bmatrix} e^{[\omega]\theta} & \left(I\theta + (1-\cos \theta) [\omega] + (\theta - \sin \theta)[\omega]^2\right)v \\ 0 & 1\end{bmatrix},
\end{equation}
where
\begin{equation}
e^{[\omega]\theta} = I + \sin \theta [\omega] + (1 -\cos \theta) [\omega]^2.
\end{equation}
The equation for $e^{[\omega]\theta} \in \text{SO}(3)$ is known as {\it Rodrigues' formula}.

For a prismatic joint, the term $e^{[\omega]\theta}$ becomes the identity matrix $I$, and $G(\theta)$ simplifies to $I\theta$. Therefore, the closed-form matrix exponential expression for the prismatic joint is given by:
\begin{equation}
e^{[\mathcal{S}]\theta} = \begin{bmatrix} I & v\theta \\ 0 & 1\end{bmatrix}.
\end{equation}

\newpage
\subsection{Implementation Details for ScrewSplat}
In this section, we describe the implementation details of ScrewSplat and its optimization process, including the initialization of core components, periodic re-initialization, selection of meaningful kinematic structures, and additional details related to the optimization process.

\textbf{Initialization of Core Components.}
For the part-aware Gaussian primitives, we initially spawn 10,000 primitives. For each primitive $\mathcal{H}_i = (T_i, s_i, \sigma_i, c_i, m_i)$, we generally adopt the same initialization scheme as used in the original Gaussian Splatting method for the variables $(T_i, s_i, \sigma_i, c_i)$. It is worth noting that the position $\mu_i$ of the Gaussian primitive from the pose $T_i = [R_i, \mu_i]$ is sampled from a uniform distribution: in the simulation environment, it is sampled from $[-1, 1]^3$, , and in real-world experiments, it is sampled from $[-0.7, 0.7]^3$. We found that initializing the positions within the camera's range results in better performance. The part probability $m_i \in \Delta^{n_s}$ is initialized as a uniform distribution (i.e., an element-wise constant vector).

For the screw primitives, we initially spawn eight revolute joint axes and eight prismatic joint axes (i.e., a total of 16 joint axes). For each screw primitive $\mathcal{A}_j = (\mathcal{S}_j, \gamma_j)$, we first sample a six-dimensional real vector from a uniform distribution over $[-0.5, 0.5]^3$ and then normalize it to satisfy the constraints for both revolute and prismatic joints. Specifically, let the sampled six-dimensional vector be $[x, q]$, where $x \in \mathbb{R}^3$ and $q \in \mathbb{R}^3$. For a revolute joint $[\omega, v]$, $\omega$ is set to $x / ||x||$ and $v$ is set to $- \omega \times q$. For a prismatic joint $[\omega, v]$, $\omega$ is set to the zero vector, and $v$ is set to $q / ||q||$. The screw confidence $\gamma_j$ is initialized to 0.9. 

For the joint angle vectors, we generate as many joint angle vector variables as there are configurations used in the observations. The joint angle $\theta_k$ is initialized to the zero vector.

\textbf{Periodic Re-initialization of Part-aware Components.}
We periodically reset all screw confidences $\gamma_j$ and part probabilities $m_i$. This periodic reset helps ScrewSplat effectively discover meaningful screw primitives. All screw confidences $\gamma_j$ and part probabilities $m_i$ are periodically reset to 0.9 and uniform distributions, respectively. We note that the reset period should be asynchronous with the original Gaussian Splatting opacity reset period for effectiveness. If the iterations are synchronized, we observe that ScrewSplat deteriorates significantly during that iteration. Additionally, at the same interval, we also re-initialize the joint angles $\theta_k$ and the poses of all part-aware Gaussian primitives $T_i$. First, we randomly select one joint angle $\theta_m$ from the set $\{\theta_1, \cdots, \theta_{n_a}\}$, and then convert all joint angles $\theta_k$ by substracting $\theta_m$, i.e., $\theta \leftarrow \theta_k - \theta_m$. For each pose $T_i$, we first select the part index $j^*$ corresponding to the highest value in the $m_i$ vector (i.e., the index $j$ with the highest $m_{ij}$ in $m_i$). Then, we convert $T_i$ as follows:
\begin{equation}
    T_i \leftarrow e^{[S_{j^*}]\theta_{mj^*}}T_i.
\end{equation}
This re-initialization of joint angles and poses has the effect of moving all Gaussian primitives to a canonical pose. Combined with the reset of screw confidences and part probabilities, this process synergistically aids in the discovery of appropriate geometries and kinematic structures.

\textbf{Selecting Meaningful Screw Primitives.}
Once ScrewSplat has converged to some extent, we eliminate meaningless screw primitives and fine-tune the ScrewSplat model using only the meaningful screws for a certain number of iterations. There are two main criteria for eliminating meaningless screw primitives. The first criterion involves screw primitives with a confidence $\gamma_j$ below a certain threshold (we set this threshold to 0.1). These primitives have little impact on rendering and are therefore considered trivially eliminated. In this case, the part-aware Gaussian primitives associated with these screw primitives are also removed. The second criterion involves screw primitives $\mathcal{A}_j$ where the difference between the maximum and minimum values in the set $\{\theta_{1j}, \cdots, \theta_{n_aj}\}$ (i.e., the interval of the joint angle bounds) is below a certain threshold (we set this threshold to 0.1 for revolute joints and 0.03 for prismatic joints). A small interval indicates that the joint angles have converged to constant values, meaning the corresponding part-aware Gaussian primitives should originally belong to the static base. In this case, we delete the screw primitive but re-initialize the Gaussian primitives to belong to the static base. Specifically, we randomly select a $\theta$ from the set $\{\theta_{1j}, \cdots, \theta_{n_aj}\}$, and then we convert $T_i$ as follows, similar to the re-initialization process described above:
\begin{equation}
    T_i \leftarrow e^{[S_{j}]\theta}T_i.
\end{equation}

\textbf{Learning Rates and Hyperparameters.}
For the variables $(T_i, s_i, \sigma_i, c_i)$, used in Gaussian Splatting, we use the same parameters as those previously employed. For the part probability $m_i$, a learning rate of 0.1 is applied before passing through the softmax. For the screw axis $\mathcal{S}_j$, a learning rate of 0.003 is used before normalization. The screw confidence $\gamma_j$ uses a learning rate of 0.01 before passing through the sigmoid. The joint angle $\theta_k$ uses a learning rate of 0.01. 

\textbf{Efficient Rendering and Backpropagation.}
To speed up the rendering and backward process during optimization, ScrewSplat only considers screw primitives with a confidence $\gamma_j$ greater than a certain threshold (we set this threshold to 0.1) for rendering the RGB.

\subsection{Details for Articulated Object Manipulation}
In this section, we describe the details of articulated object manipulation, including additional information on controlling joint angles with text prompts using ScrewSplat, as discussed in Section 4.2, details on the current state estimation step mentioned in Section 5.2, and a brief explanation of robot trajectory planning.

\textbf{Controlling Joint Angles Using ScrewSplat.}
For the CLIP model, we use \texttt{openai/clip-vit-base-patch32} from Huggingface. For Bayesian optimization, We use the \texttt{gp\_minimize} function from the \texttt{scikit-optimize} library to optimize the joint angles. The optimization is guided by the Expected Improvement (EI) acquisition function and is performed over 50 function evaluations, with the first 10 being random joint angle samples. For recognition, we calculate the element-wise min and max of the optimized joint angles $\theta_k$ to set the joint limits, and the search space is defined based on these joint limits. The rationale behind using directional CLIP loss instead of simple cosine similarity loss, and Bayesian optimization instead of gradient-based optimization, is discussed in detail in the Appendix D.3.

{\bf Current State Estimation Step.}
As discussed in Section 5.2, for a changed articulated object configuration after recognition, we additionally estimate the object’s current joint angle. To achieve this, we first obtain additional RGB observations of the articulated object, and then minimize the rendering loss $\mathcal{L}_render$ used in Gaussian Splatting to find the current joint angle:
\begin{equation}
\mathcal{L}_{\text{estimate}} = \mathcal{L}_{\text{render}}.    
\end{equation}
For optimization, we also use Bayesian optimization, and the parameters employed are the same as those used in the text-guided object manipulation described above.

{\bf Robot Trajectory Planning for Real-world Articulated Object Manipulation.}
Given the current joint angle $\theta_c \in \mathbb{R}$ and target joint angle $\theta_t \in \mathbb{R}$, along with a screw primitive $\mathcal{A}_j = (\mathcal{S}_j, \gamma_j)$, we propose a simple robot trajectory planning approach for real-world articulated object manipulation. The trajectory planning consists of two main stages: first, planning the robot gripper's tip trajectory, and second, planning the gripper's $\text{SE}(3)$ trajectory based on the tip trajectory. 

To plan the tip's trajectory, we first identify an affordance point. This involves collecting the centers $\mu_i$ of valid part-aware Gaussian primitives $\mathcal{H}_i$. For revolute joints, we select a subset of centers that lie within the top 10th percentile, the farthest from the axis. For prismatic joints, we select the subset of centers closest to the robot base, within the 20th percentile along the axis dimension. From this subset, the center of the cluster is selected as the affordance point. Subsequently, the tip trajectory is generated by moving the affordance point from $\theta_c - \theta_o$, where $\theta_o$ is an offset designed to help avoid object-robot collisions, to $\theta_t$ along the screw axis $\mathcal{S}_j$. 

After designing the tip trajectory, we then plan the gripper's $\text{SE}(3)$ trajectory, ensuring that the gripper's tip follows the tip trajectory while maintaining a fixed orientation. Typically, we set the orientation so that the robot gripper faces toward the ground. Once the $\text{SE}(3)$ trajectory is obtained, we solve the inverse kinematics to compute the final robot joint angle trajectory for articulated object manipulation.

\subsection{Mixed-integer Optimization Formulation vs. ScrewSplat Formulation}
In this section, we briefly compare the original mixed-integer optimization formulation for articulated object recognition with our relaxed formulation using ScrewSplat. As mentioned in the introduction, articulated object recognition is particularly challenging due to its mixed-integer optimization structure, which involves both continuous variables (e.g., 3D geometry and joint angles) and discrete, combinatorial variables (e.g., part segmentation labels, joint types, and joint counts). Formally, the general mixed-integer optimization formulation can be expressed as:
\begin{align*}
    \min_{\{\mathcal{H}_i\}, \textcolor{red}{\{\mathcal{S}_j\}}, \{\theta_k\}, \textcolor{red}{n_s}} \quad & \sum_k \mathcal{L}_{\text{render}}\big(\textcolor{red}{\pi_{\text{GS}}}(\theta_k; \{\mathcal{H}_i\}, \textcolor{red}{\{\mathcal{S}_j\}}), I_{\text{gt}}\big) + \textcolor{red}{\beta n_s}\\
    \text{subject to} \quad & \mathcal{H}_i = (T_i, s_i, \sigma_i, c_i, m_i), \quad \textcolor{red}{m_i \in \{0, 1, \ldots, n_s\}}, \quad i = 1, \dots, n_g,  \\
    & \mathcal{S}_j \in \mathbb{R}^6, \quad j = 1, \dots, n_s, \\
    & \theta_k \in \mathbb{R}^{n_s}, \quad k = 1, \dots, n_a, \\
    & \textcolor{red}{n_s \in \mathbb{N}},
\end{align*}
where
\begin{itemize}
    \item $\mathcal{H}_i$ denotes a part-aware Gaussian primitive, where $(T_i, s_i, \sigma_i, c_i)$ are the parameters of the corresponding 3D Gaussian, and $m_i$ is its segmentation label.
    \item $\mathcal{S}_j$ represents the screw axis of the $j$-th movable part in the articulated object.
    \item $\theta_k = (\theta_{k1}, \dots, \theta_{kn_s}) \in \mathbb{R}^{n_s}$ is the joint angle vector for the $k$-th configuration of the articulated object.
    \item $n_s$ is the number of screws, corresponding to the number of movable parts.
    \item $\pi_{\text{GS}}$ is the standard rendering function used in Gaussian Splatting; during rendering, the $i$-th part-aware Gaussian primitive is transformed into a standard Gaussian with parameters $\big(e^{[\mathcal{S}_{m_i}] \theta_{k m_i}} T_i, s_i, \sigma_i, c_i\big)$.
\end{itemize}
Several previous works simplify this problem by making certain assumptions. For example, some assume a known number of articulated components (i.e., $n_s$ is no longer treated as an optimization variable), while others use intermediate procedures such as point correspondence matching or part clustering to obtain part segmentation labels a priori (i.e., $m_i$ is no longer treated as an optimization variable).

Our goal is to address this optimization problem without relying on intermediate steps, auxiliary data, or prior knowledge of joint types or counts. Rather than solving the mixed-integer optimization problem directly, we reformulate it into a differentiable relaxed version (as described in Section 4), which enables effective optimization. Our formulation can be expressed as:
\begin{align*}
    \min_{\{\mathcal{H}_i\}, \textcolor{blue}{\{\mathcal{A}_j\}}, \{\theta_k\}} \quad & \sum_k \mathcal{L}_{\text{render}} \big(\textcolor{blue}{\pi_{\text{SS}}}(\theta_k; \{\mathcal{H}_i\}, \textcolor{blue}{\{\mathcal{A}_j\}}), I_{\text{gt}}\big) + \textcolor{blue}{\beta \sum_{j} \sqrt{\gamma_j}} \\
    \text{subject to} \quad & \mathcal{H}_i = (T_i, s_i, \sigma_i, c_i, m_i), \quad \textcolor{blue}{m_i \in \Delta^{n_s}}, \quad i = 1, \dots, n_g,  \\
    & \textcolor{blue}{\mathcal{A}_j = (\mathcal{S}_j, \gamma_j)}, \quad \mathcal{S}_j \in \mathbb{R}^6, \quad \textcolor{blue}{\gamma_j \in [0, 1]}, \quad j = 1, \dots, n_s, \\
    & \theta_k \in \mathbb{R}^{n_s}, \quad k = 1, \dots, n_a,
\end{align*}
where
\begin{itemize}
    \item $\mathcal{H}_i$ denotes a part-aware Gaussian primitive, where $(T_i, s_i, \sigma_i, c_i)$ are the parameters of the corresponding 3D Gaussian, and \textcolor{blue}{$m_i$ is a probability simplex over the movable parts}.
    \item $\mathcal{A}_j$ represents the screw primitive of the $j$-th movable part, where $\mathcal{S}_j$ is the screw axis and \textcolor{blue}{$\gamma_j$ denotes its confidence}.
    \item $\theta_k = (\theta_{k1}, \dots, \theta_{kn_s}) \in \mathbb{R}^{n_s}$ is the joint angle vector for the $k$-th configuration of the articulated object.
    \item \textcolor{blue}{$\pi_{\text{SS}}$ is the rendering function of ScrewSplat}, as described in Section 4.1 (``RGB Rendering Procedure with ScrewSplat'').
\end{itemize}

%% file: 00appendix/A3experiment.tex

\newpage
\section{Experimental Details}

\subsection{Additional Details for Recognition Experiments}
In this section, we provide further details on the evaluation dataset, baseline implementations, and evaluation metrics.

\textbf{Dataset.} 
We first select ten single-joint objects and three multi-joint objects from distinct categories in the PartNet-Mobility dataset~\cite{xiang2020sapien}. For the camera parameters, we use the same intrinsic parameters and resolution as those of the Intel RealSense D435, and sample 48 camera positions uniformly distributed over a hemisphere of radius 1, centered on each articulated object. The camera orientations are set such that they face the center of the object. To ensure that the objects are fully visible, we rescale each object.

For each articulated object, we generate observation data for five different configurations (i.e., joint angles). For the single-joint objects, we use evenly spaced joint angles within the joint limits, while for the multi-joint objects, we randomly sample joint angle vectors. For each object and configuration, we render both RGB and depth images (which are used for optimizing PARIS* and DTA) from the 48 camera poses. We use Blender~\cite{blender} to render the RGB and depth images.

{\bf Baseline Methods.} 
For brief description of baselines including {\it PARIS}~\cite{liu2023paris} and {\it DTA}~\cite{weng2024neural}, we refer to Appendix A.3. It is important to note that for {\it PARIS*}, which also incorporates depth information, we use an additional depth rendering loss with depth supervision during the optimization step. These baseline methods currently accept input data for only two object configurations, so we use only the observations corresponding to two of the five configurations generated in our dataset. For the single-joint objects, we compare against PARIS, PARIS*, and DTA, and we select the two middle configurations from the evenly spaced set (specifically, the observation data from the 2nd and 4th configurations are used). For the multi-joint objects, we compare only against DTA, and since the five configurations are randomly sampled, we randomly select two configurations for use.

{\it ScrewSpawn} is an ablation model that follows the ScrewSplat framework but spawns only a single screw (with a known joint type). Specifically, in ScrewSpawn, only one screw primitive, $\mathcal{A}_1 = (\mathcal{S}_1, \gamma_1)$, is spawned, and the confidence value $\gamma_1$ is fixed at 1.0 (i.e., it is not treated as an optimization variable). Thus, optimization is performed only over the screw axis variable $\mathcal{S}_1$ among the variables of $\mathcal{A}_1$. Since there is only a single screw, the parsimony loss is also not applied. All other optimization details follow those of ScrewSplat. We use all observations corresponding to the five configurations generated in our dataset for optimization.

{\bf Evaluation Metrics.}
We again note that we adopt three types of metrics, including {\it geometry}, {\it motion}, and {\it appearance}, for articulated object recognition experiment. For {\it geometry} metric, we first reconstruct meshes from the recognized models. Specifically, for PARIS, PARIS*, and DTA, we use the marching cubes method~\cite{lorensen1998marching}. For ScrewSpawn and ScrewSplat, we render depth images from multiple designated camera views and fuse them using the Truncated Signed Distance Function (TSDF). The final mesh is then extracted using the marching cubes method from the TSDF. From each reconstructed mesh, we uniformly sample 2,048 points to obtain a point cloud and compute the bi-directional Chamfer-$l_2$ distances as described in Section 5.1.

For {\it motion} metric, the calculation for single-joint objects follows the procedure described in Section 5.1. For multi-joint objects, given $n$ ground-truth screw axes and $m$ recognized screw axes, we first perform bipartite matching between the two sets based on angular error and axis position error. Then, for each of the $n$ ground-truth axes, we compute the motion metrics accordingly.

For {\it appearance} metric, we use the Peak Signal-to-Noise Ratio (PSNR) and Structural Similarity Index Measure (SSIM), computed on rendered images under unseen joint angles. These unseen joint angles correspond to the midpoint values between the joint angles of the object configurations used during optimization. At these intermediate joint angles, we render RGB images from 48 camera views, following the same procedure described in the dataset section, and use them as ground-truth images. We then compute the PSNR and SSIM scores between the rendered outputs of the recognized models and these ground-truth images. Since DTA does not render RGB images and focuses solely on estimating geometry and kinematic structure, we exclude it from the appearance metric evaluation.

{\bf Computational Aspect.}
Optimizing ScrewSplat over 30,000 iterations takes approximately 7-9 minutes on average, depending on the object, measured on a GeForce RTX 4090 GPU. On the same hardware, PARIS and DTA take about 3 and 15 minutes on average, respectively. ScrewSplat occupies approximately 30MB per object, with peak GPU memory usage of approximately 2.8GB, as measured by \texttt{torch.cuda.max\_memory\_allocated}. 

\subsection{Additional Details for Real-world Manipulation Experiments}

We use the 7-DoF Franka Emika Panda robot equipped with a parallel-jaw gripper and an Intel RealSense D435 camera mounted on the gripper. We sample 24 camera positions uniformly distributed over a {\it partial} hemisphere of radius 0.85 (i.e., providing only a partial view), where the camera orientations are set to face the center of the workspace. Among these, we use 16 camera poses for which the robot has valid inverse kinematics (IK) solutions, as the camera is mounted on the robot arm.

Multi-view RGB observations are then collected under five object joint configurations, which are manually set by a human operator. The collected RGB images are processed into masked object images using the pretrained segmentation network SAM~\cite{kirillov2023segany}. These masked images are used as input for recognition with ScrewSplat. Additionally, in the real-world experiment, we set the weight of the parsimony loss to 0.005. The same 16 camera poses are also used when estimating the current object state.

%% file: 00appendix/A4additional.tex

\newpage
\section{Additional Experimental Results}

\subsection{Additional Results for Articulated Object Recognition Experiments}

\textbf{Single-joint Objects.}
Figure~\ref{fig:supple_recognition_results} presents additional examples of single-joint articulated object recognition. The overall trend is similar to that shown in Figure 5 of the main text. While PARIS and PARIS* successfully recognize the geometry and kinematic structure of some objects, they also exhibit several failure cases. DTA generally succeeds in recognizing all the additional objects. ScrewSpawn fails to recognize all objects except the scissor. ScrewSplat, on the other hand, successfully recognizes all objects and predicts more accurate and precise geometry, as well as more accurate kinematic structures, compared to DTA.

Table~\ref{tab:single_joint_detail} reports the object-wise quantitative recognition results. We demonstrate that ScrewSplat generally achieves the highest overall performance across geometry, motion, and visual appearance. In particular, compared to the existing baselines, ScrewSplat consistently outperforms all others in predicting the geometry of the movable parts and joint axes. Although ScrewSpawn outperforms ScrewSplat for specific objects such as the scissor, we additionally observe that optimization often falls into local minima for most other objects.

\begin{wrapfigure}{r}{0.5\textwidth}
    \centering
    \vspace{-15pt}
    \includegraphics[width=1\linewidth]{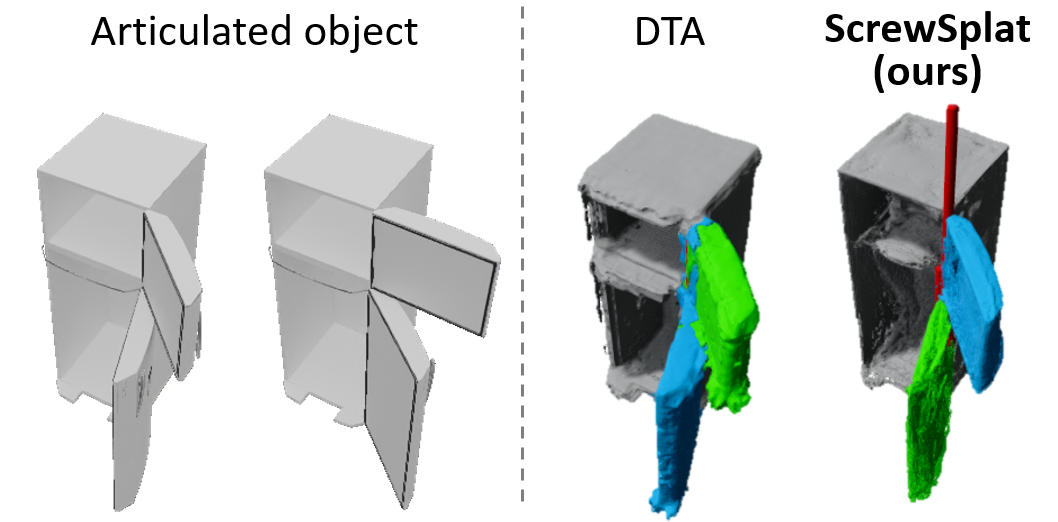}
    \caption{Additional results showing reconstructed meshes and screw axes for a multi-joint object using each method.}
    \vspace{-15pt}
    \label{fig:supple_recognition_multi_results}
\end{wrapfigure}
\textbf{Multi-joint Objects.}
Figure~\ref{fig:supple_recognition_multi_results} presents an additional example of multi-joint articulated object recognition. In this object, DTA also achieves a reasonably accurate recognition. Along with Figure 6, ScrewSplat demonstrates superior performance over DTA by predicting more precise geometry and more accurate joint axes. Table~\ref{tab:multi_joint_detail} shows the object-wise quantitative results for multi-joint articulated object recognition. From the geometry perspective, while DTA slightly outperforms ScrewSplat in predicting the geometry of the static and whole parts, ScrewSplat significantly outperforms DTA in predicting the geometry of the movable parts. From the kinematic structure perspective, ScrewSplat generally shows better performance. In particular, for the 3-joint object, DTA completely fails by predicting an incorrect joint axis, whereas ScrewSplat successfully identifies the correct one.


\newpage
\begin{figure}[h]
    \centering
    \includegraphics[width=\linewidth]{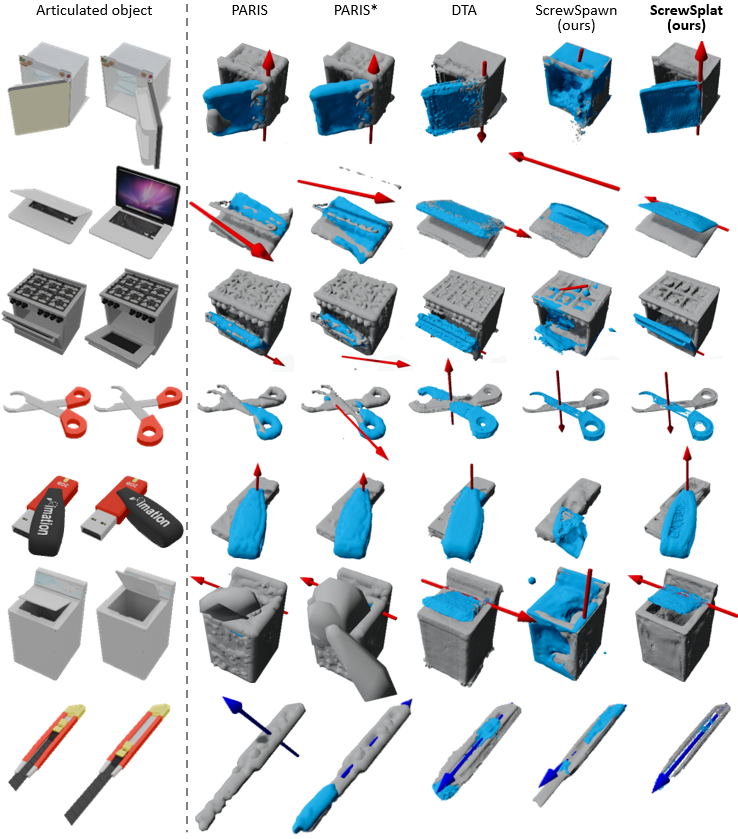}
    \vspace{-15pt}
    \caption{Additional results showing reconstructed meshes and screw axes for all single-joint objects used in the experiment (excluding the folding chair, storage box, and stapler shown in Figure 5), using each method. Static parts are shown in gray, movable parts in cyan, revolute joints in red, and prismatic joints in blue.}
    \vspace{-5pt}
    \label{fig:supple_recognition_results}
\end{figure}

\newpage
\begin{table}[h]
\footnotesize
\centering
\caption{Object-wise recognition performance for single-joint objects.}
\label{tab:single_joint_detail}
\begin{tabular}{llccccccccccc}
\multicolumn{2}{c}{} &\multicolumn{3}{c}{Geometry ($\downarrow$)} &\multicolumn{1}{c}{} &\multicolumn{2}{c}{Motion ($\downarrow$)} &\multicolumn{1}{c}{} &\multicolumn{2}{c}{Appearance ($\uparrow$)} \\
\cline{3-5}
\cline{7-8} 
\cline{10-11} 

\multicolumn{1}{c}{\bf OBJECT} &\multicolumn{1}{c}{\bf METHOD}  &\multicolumn{1}{c}{CD-s} &\multicolumn{1}{c}{CD-m} &\multicolumn{1}{c}{CD-w} &\multicolumn{1}{c}{} &\multicolumn{1}{c}{Ang.} &\multicolumn{1}{c}{Pos.} &\multicolumn{1}{c}{} &\multicolumn{1}{c}{PSNR} &\multicolumn{1}{c}{SSIM} \\
\hline 

\multirow{5}{*}{FoldChair} 
& PARIS~\cite{liu2023paris}        & 8.285 & 0.223 & 6.359 &   & 2.472 & 0.277 &  & 26.831 & 0.970 \\
& PARIS*~\cite{liu2023paris}        & 3.244 & 0.206 & 1.756 &   & 1.302 & 0.105 &  & 28.558 & 0.973 \\
& DTA~\cite{weng2024neural}       & 0.551 & 0.224 & 0.367 &  & 0.262 & 0.037 &  & - & - \\
& ScrewSpawn  & 0.700 & 0.220 & 0.170 & & 0.498 & 0.013 & & 29.440 & 0.982 \\
& ScrewSplat  & \textbf{0.052} & \textbf{0.051} & \textbf{0.090} & & \textbf{0.058} & \textbf{0.017} & & \textbf{32.970} & \textbf{0.991} \\
\hline

\multirow{5}{*}{Fridge} 
& PARIS~\cite{liu2023paris}        & 6.746 & 0.324 & 4.365 &   & 1.722 & 1.501 &  & 34.084 & 0.981 \\
& PARIS*~\cite{liu2023paris}        & 4.341 & 0.304 & 2.070 &   & 2.252 & 0.900 &  & 33.927 & 0.981 \\
& DTA~\cite{weng2024neural}       & 0.469 & 0.221 & 0.358 &  & 0.287 & 0.024 &  & - & - \\
& ScrewSpawn & 0.635 & 15.428 & 1.881 & & 27.423 & 0.111 & & 27.590 & 0.984 \\
& ScrewSplat & \textbf{0.256} & \textbf{0.117} & \textbf{0.289} & & \textbf{0.231} & \textbf{0.004} & & \textbf{40.110} & \textbf{0.995} \\
\hline

\multirow{5}{*}{Laptop} 
& PARIS~\cite{liu2023paris}        & 239.870 & 113.542 & 192.582 &   & 22.119 & 1.246 &  & 24.790 & 0.958 \\
& PARIS*~\cite{liu2023paris}        & 293.719 & 19.709 & 194.828 &   & 13.482 & 2.149 &  & 24.557 & 0.957 \\
& DTA~\cite{weng2024neural}       & 0.926 & 0.541 & \textbf{0.341} &  & 0.227 & 0.166 &  & - & - \\
& ScrewSpawn  & 0.076 & 0.211 & 0.209 & & 0.062 & 2.375 & & 27.070 & 0.977 \\
& ScrewSplat & \textbf{0.322} & \textbf{0.170} & 0.347 & & \textbf{0.071} & \textbf{0.015} & & \textbf{38.260} & \textbf{0.994} \\
\hline

\multirow{5}{*}{Oven} 
& PARIS~\cite{liu2023paris}        & 15.400 & 9.209 & 9.332 &   & 6.547 & 4.285 &  & 29.170 & 0.952 \\
& PARIS*~\cite{liu2023paris}        & 21.212 & 11.606 & 16.419 &   & 31.119 & 2.760 &  & 27.305 & 0.947 \\
& DTA~\cite{weng2024neural}       & \textbf{0.561} & 0.242 & \textbf{0.512} &  & 0.251 & 0.097 &  & - & - \\
& ScrewSpawn  & 0.834 & 23.083 & 1.223 & & 33.489 & 1.451 & & 26.650 & 0.968 \\
& ScrewSplat  & 0.617 & \textbf{0.204} & 0.536 & & \textbf{0.125} & \textbf{0.007} & & \textbf{35.010} & \textbf{0.983} \\
\hline

\multirow{5}{*}{Scissor} 
& PARIS~\cite{liu2023paris}        & 3.630 & 0.675 & 0.198 &   & 19.904 & 0.994 &  & 29.972 & 0.975 \\
& PARIS*~\cite{liu2023paris}        & 7.625 & 1.438 & 2.195 &   & 59.327 & 0.896 &  & 28.656 & 0.971 \\
& DTA~\cite{weng2024neural}       & 0.337 & 0.299 & 0.339 &  & 0.136 & 0.029 &  & - & - \\
& ScrewSpawn  & \textbf{0.047} & \textbf{0.046} & \textbf{0.068} & & \textbf{0.099} & \textbf{0.004} & & \textbf{39.770} & \textbf{0.997} \\
& ScrewSplat  & \textbf{0.047} & 0.054 & 0.070 & & 0.109 & 0.016 & & 38.990 & 0.996 \\
\hline

\multirow{5}{*}{Stapler} 
& PARIS~\cite{liu2023paris}        & 151.146 & 6.510 & 85.308 &   & 4.426 & 0.064 &  & 21.053 & 0.954 \\
& PARIS*~\cite{liu2023paris}        & 99.810 & 20.018 & 61.348 &   & 5.205 & 2.306 &  & 21.198 & 0.954 \\
& DTA~\cite{weng2024neural}       & 0.320 & 1.167 & 0.181 &  & 0.222 & 2.058 &  & - & - \\
& ScrewSpawn  & 0.139 & 3.256 & 0.320 & & 1.195 & 2.253 & & 21.940 & 0.975 \\
& ScrewSplat  & \textbf{0.127} & \textbf{0.685} & \textbf{0.126} & & \textbf{0.054} & \textbf{0.005} & & \textbf{36.850} & \textbf{0.995} \\
\hline

\multirow{5}{*}{USB} 
& PARIS~\cite{liu2023paris}        & 0.200 & 0.236 & 0.215 &   & 0.688 & 3.502 &  & 28.068 & 0.973 \\
& PARIS*~\cite{liu2023paris}        & 0.207 & 0.228 & 0.200 &   & 0.989 & 0.048 &  & 26.999 & 0.970 \\
& DTA~\cite{weng2024neural}       & 0.586 & 0.369 & 0.288 &  & 0.172 & 0.023 &  & - & - \\
& ScrewSpawn & 0.404 & 5.137 & 1.543 & & 3.159 & 0.168 & & 22.220 & 0.969 \\
& ScrewSplat  & \textbf{0.236} & \textbf{0.105} & \textbf{0.234} & & \textbf{0.047} & \textbf{0.001} & & \textbf{35.300} & \textbf{0.993} \\
\hline

\multirow{5}{*}{Washer} 
& PARIS~\cite{liu2023paris}        & 95.739 & 8.080 & 89.933 &   & 16.599 & 4.287 &  & 32.424 & 0.981 \\
& PARIS*~\cite{liu2023paris}        & 54.112 & 0.625 & 49.712 &   & 5.279 & 4.778 &  & 35.970 & 0.986 \\
& DTA~\cite{weng2024neural}       & \textbf{0.435} & 0.527 & \textbf{0.447} &  & 0.397 & 0.026 &  & - & - \\
& ScrewSpawn  & 1.197 & 29.681 & 0.781 & & 88.232 & 0.844 & & 33.890 & 0.992 \\
& ScrewSplat  & 0.714 & \textbf{0.335} & 0.449 & & \textbf{0.079} & \textbf{0.014} & & \textbf{41.510} & \textbf{0.996} \\
\hline

\multirow{5}{*}{Knife}
& PARIS~\cite{liu2023paris}        & 7.438 & F & 4.554 &   & 61.590 & - &  & 30.554 & 0.988 \\
& PARIS*~\cite{liu2023paris}        & 3.611 & 32.524 & 3.663 &   & 1.879 & - &  & 30.161 & 0.988 \\
& DTA~\cite{weng2024neural}       & 0.355 & 0.410 & 0.359 &  & 0.047 & - &  & - & - \\
& ScrewSpawn  & 1.088 & 28.176 & 2.439 & & 5.993 & - & & 31.230 & 0.994 \\
& ScrewSplat  & \textbf{0.039} & \textbf{0.038} & \textbf{0.046} & & \textbf{0.031} & & & \textbf{41.090} & \textbf{0.998} \\
\hline

\multirow{5}{*}{Storage} 
& PARIS~\cite{liu2023paris}        & 11.698 & 23.487 & 9.072 &   & 40.492 & - &  & 29.482 & 0.968 \\
& PARIS*~\cite{liu2023paris}        & 9.181 & 25.644 & 6.367 &   & 42.036 & - &  & 29.220 & 0.968 \\
& DTA~\cite{weng2024neural}       & 0.842 & 1.281 & \textbf{0.407} &  & 2.372 & - &  & - & - \\
& ScrewSpawn & 1.056 & 10.426 & 0.823 & & 88.541 & - & & 31.340 & 0.983 \\
& ScrewSplat & \textbf{0.783} & \textbf{0.355} & 0.421 & & \textbf{0.030} & - & & \textbf{40.650} & \textbf{0.992} \\
\hline

\end{tabular}
\end{table}


\begin{sidewaystable}[]
\footnotesize
\centering
\caption{Object-wise recognition performance for multi-joint objects.}
\label{tab:multi_joint_detail}
\begin{tabular}{lcccccccccccccccccccc}
\multicolumn{2}{c}{} &\multicolumn{5}{c}{Geometry ($\downarrow$)} &\multicolumn{1}{c}{} &\multicolumn{6}{c}{Motion ($\downarrow$)} &\multicolumn{1}{c}{} &\multicolumn{2}{c}{Appearance ($\uparrow$)} \\
\cline{3-7}
\cline{9-14} 
\cline{16-17} 

\multicolumn{1}{c}{\bf OBJECT} &\multicolumn{1}{c}{\bf METHOD}  &\multicolumn{1}{c}{CD-s} &\multicolumn{1}{c}{CD-m$_1$} &\multicolumn{1}{c}{CD-m$_2$} &\multicolumn{1}{c}{CD-m$_3$} &\multicolumn{1}{c}{CD-w} &\multicolumn{1}{c}{} &\multicolumn{1}{c}{Ang$_1$.} &\multicolumn{1}{c}{Pos$_1$.} &\multicolumn{1}{c}{Ang$_2$.} &\multicolumn{1}{c}{Pos$_2$.} &\multicolumn{1}{c}{Ang$_3$.} &\multicolumn{1}{c}{Pos$_3$.} &\multicolumn{1}{c}{} &\multicolumn{1}{c}{PSNR} &\multicolumn{1}{c}{SSIM} \\
\hline 

\multirow{2}{*}{Fridge-2} 
& DTA~\cite{weng2024neural} & \textbf{0.402} & 5.886 & 0.466 & - & \textbf{0.424} & & 0.733 & 0.005 & 0.733 & 0.005 & - & - & & - & -\\
& ScrewSplat                & 0.517 & \textbf{0.057} & \textbf{0.093} & - & 0.531 &  & \textbf{0.120} & \textbf{0.001} & \textbf{0.091} & \textbf{0.003} & - & - & & 36.65 & 0.990\\
\hline

\multirow{2}{*}{Storage-2} 
& DTA~\cite{weng2024neural} & \textbf{0.524} & 3.604 & 0.269 & - & \textbf{0.500} & & \textbf{0.114} & 0.114 & 0.480 & - & - & - & & - & -\\
& ScrewSplat                & 1.033 & \textbf{0.303} & \textbf{0.083} & - & 1.009 &  & 0.136 & \textbf{0.000} & \textbf{0.109} & - & - & - & & 37.20 & 0.988\\
\hline

\multirow{2}{*}{Storage-3} 
& DTA~\cite{weng2024neural} & 0.790 & 2.984 & 7.211 & 25.308 & 0.466 & & 1.017 & 0.084 & 12.474 & 0.220 & 48.522 & 259.235 & & - & -\\
& ScrewSplat                & \textbf{0.476} & \textbf{0.044} & \textbf{0.081} & \textbf{0.068} & \textbf{0.459} &  & \textbf{0.074} & \textbf{0.003} & \textbf{0.123} & \textbf{0.001} & \textbf{0.173} & \textbf{0.002} & & 36.43 & 0.983\\
\hline

\end{tabular}
\end{sidewaystable}

\newpage
\subsection{Recognition Performance of ScrewSplat from Two-Configuration Observations}
When optimizing ScrewSplat, we use RGB observations collected from five different configurations of the articulated object, whereas the other baselines only take observations from two configurations as input. In fact, to put it differently, the other baselines cannot handle observations from an arbitrary number of configurations, while ScrewSplat can. This makes ScrewSplat a more flexible algorithm capable of processing a richer set of information at once. To verify whether ScrewSplat still performs well when given limited input (i.e., observations from limited configurations), we conduct an ablation study using only two configurations as input.

Table~\ref{tab:two_config} shows the recognition performance of ScrewSplat optimized using observations from two and five object configurations. Before analysis, we confirm that the performance was slightly sensitive to the weight of the parsimony loss when only using two configurations. Therefore, we optimize ScrewSplat over 10 different weights, ranging from 0.001 to 0.010, and evaluate the model with the best performance. From the table, it can be seen that there is little performance difference for most objects. In particular, for examples like the laptop and scissor, using only two configurations actually yields better performance in terms of geometry and motion metrics. However, even with the best parsimony loss weight set for two configurations, recognition failed for specific objects like the stapler and USB. In conclusion, ScrewSplat can recognize objects to some extent using observations from two configurations of articulated objects, and as discussed in the limitations and future directions, improvements in optimization techniques could lead to better performance in these cases.

\begin{table}[]
\footnotesize
\centering
\caption{Recognition performance of ScrewSplat optimized using observations from two and five object configurations.}
\label{tab:two_config}
\begin{tabular}{llccccccccccc}
\multicolumn{2}{c}{} &\multicolumn{3}{c}{Geometry ($\downarrow$)} &\multicolumn{1}{c}{} &\multicolumn{2}{c}{Motion ($\downarrow$)} &\multicolumn{1}{c}{} &\multicolumn{2}{c}{Appearance ($\uparrow$)} \\
\cline{3-5}
\cline{7-8} 
\cline{10-11} 
\multicolumn{1}{c}{\bf OBJECT} &\multicolumn{1}{c}{\bf METHOD}  &\multicolumn{1}{c}{CD-s} &\multicolumn{1}{c}{CD-m} &\multicolumn{1}{c}{CD-w} &\multicolumn{1}{c}{} &\multicolumn{1}{c}{Ang.} &\multicolumn{1}{c}{Pos.} &\multicolumn{1}{c}{} &\multicolumn{1}{c}{PSNR} &\multicolumn{1}{c}{SSIM} \\
\hline

\multirow{2}{*}{FoldChair} 
& 2-config & 0.070 & 0.066 & 0.094 & & 0.212 & \textbf{0.006} & & \textbf{34.39} & \textbf{0.993} \\
& 5-config & \textbf{0.054} & \textbf{0.061} & \textbf{0.091} & & \textbf{0.058} & 0.017 & & 33.44 & 0.992 \\
\hline

\multirow{2}{*}{Fridge} 
& 2-config & 0.302 & 0.236 & 0.308 & & \textbf{0.153} & 0.007 & & 39.34 & 0.993 \\
& 5-config & \textbf{0.253} & \textbf{0.108} & \textbf{0.278} & & 0.231 & \textbf{0.004} & & \textbf{41.09} & \textbf{0.995} \\
\hline

\multirow{2}{*}{Laptop} 
& 2-config & \textbf{0.069} & \textbf{0.081} & \textbf{0.097} & & \textbf{0.066} & \textbf{0.003} & & 37.13 & 0.993 \\
& 5-config & 0.322 & 0.127 & 0.347 & & 0.071 & 0.015 & & \textbf{39.01} & \textbf{0.994} \\
\hline

\multirow{2}{*}{Oven} 
& 2-config & \textbf{0.563} & 0.474 & \textbf{0.595} & & \textbf{0.033} & 0.013 & & \textbf{35.24} & \textbf{0.983} \\
& 5-config & 0.607 & \textbf{0.274} & 0.619 & & 0.125 & \textbf{0.007} & & 34.65 & 0.982 \\
\hline

\multirow{2}{*}{Scissor} 
& 2-config & 1.915 & \textbf{0.054} & \textbf{0.064} & & \textbf{0.045} & \textbf{0.008} & & 36.66 & 0.995 \\
& 5-config & \textbf{0.047} & 0.055 & 0.067 & & 0.109 & 0.016 & & \textbf{39.46} & \textbf{0.997} \\
\hline

\multirow{2}{*}{Stapler} 
& 2-config & 0.154 & 6.199 & 1.195 & & F & 1.519 & & 21.46 & 0.971 \\
& 5-config & \textbf{0.122} & \textbf{0.577} & \textbf{0.127} & & \textbf{0.054} & \textbf{0.005} & & \textbf{36.54} & \textbf{0.995} \\
\hline

\multirow{2}{*}{USB} 
& 2-config & 0.276 & 0.151 & 0.262 & & 0.084 & 0.104 & & 28.08 & 0.979 \\
& 5-config & \textbf{0.237} & \textbf{0.106} & \textbf{0.225} & & \textbf{0.047} & \textbf{0.001} & & \textbf{36.97} & \textbf{0.994} \\
\hline

\multirow{2}{*}{Washer} 
& 2-config & 0.824 & 0.132 & 0.662 & & 0.208 & 0.028 & & 41.08 & \textbf{0.996} \\
& 5-config & \textbf{0.717} & \textbf{0.092} & \textbf{0.617} & & \textbf{0.079} & \textbf{0.014} & & \textbf{41.86} & \textbf{0.996} \\
\hline

\multirow{2}{*}{Knife} 
& 2-config & 0.044 & 0.070 & 0.048 & & 0.045 & - & & 41.14 & \textbf{0.998} \\
& 5-config & \textbf{0.039} & \textbf{0.038} & \textbf{0.047} & & \textbf{0.031} & - & & \textbf{41.23} & \textbf{0.998} \\
\hline

\multirow{2}{*}{Storage} 
& 2-config & 0.811 & 0.919 & 0.409 & & 0.180 & - & & 40.39 & 0.991 \\
& 5-config & \textbf{0.784} & \textbf{0.343} & \textbf{0.393} & & \textbf{0.030} & - & & \textbf{40.49} & \textbf{0.992} \\
\hline
\end{tabular}
\end{table}

\newpage
\subsection{Why We Use Bayesian Optimization with Directional CLIP Loss}

\begin{figure}[]
    \centering
    \includegraphics[width=\linewidth]{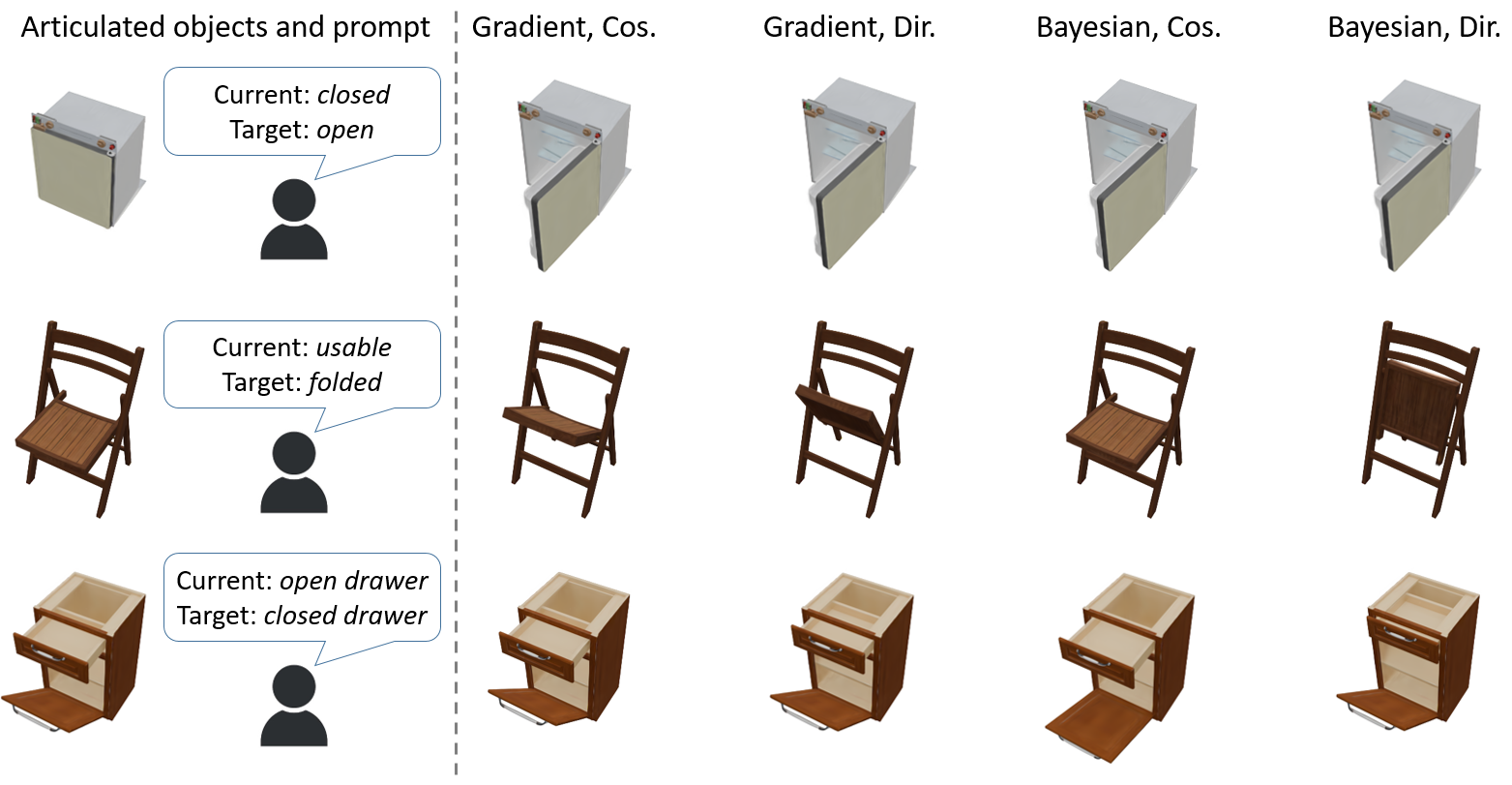}
    \caption{Comparison of directional CLIP loss versus cosine similarity loss, and Bayesian optimization versus gradient-based optimization for text-guided articulated object manipulation.}
    \label{fig:supple_sim_manipulation_results}
\end{figure}

As described in Section 4.2, we optimize the target joint angles for text-guided articulated object manipulation using directional CLIP loss~\cite{gal2022stylegan} through Bayesian optimization. However, there are more trivial alternatives. For instance, one could use a simple cosine similarity loss defined as follows~\cite{radford2021learning}:
\begin{equation}
    \mathcal{L}_{\text{CLIP-sim}} = 1 - e_I(\pi(\theta)) \cdot e_T(t_p),
\end{equation}
where the notations are consistent with those used in Section 4.2. Furthermore, since our rendering function $\pi$ is differentiable, this would allow the use of gradient-based optimization~\cite{liu2024differentiable}. In this section, we compare directional CLIP loss versus cosine similarity loss, and Bayesian optimization versus gradient-based optimization. We provide a qualitative comparison of the performance for text-guided manipulation.

Figure~\ref{fig:supple_sim_manipulation_results} shows the visual appearance of articulated objects based on the optimized joint angles for each case. Overall, the refrigerator successfully reaches the target in all cases, while for other objects, we observe partial success or failure in cases where Bayesian optimization and directional CLIP loss are not used. In more detail, we can see that the directional CLIP loss leads to joint angles that align better with the target prompt than cosine similarity loss. This suggests that, for articulated objects, considering relative directions through a text description of the current state is more effective. Even when minimizing the directional CLIP loss, we observe partial success when using gradient-based methods. This implies that the loss landscape is complex and may lead to local minima, suggesting the need for a broader search space for joint angles. Currently, with joint angles having a maximum dimension of three, Bayesian optimization works efficiently. As the dimensionality increases, there is a need to design more appropriate optimization schemes.